\documentclass[twoside,11pt]{article}

\usepackage[preprint]{jmlr2e}

\usepackage{amsmath}
\usepackage{amssymb}
\usepackage{amsfonts}
\usepackage{enumitem}
\usepackage{bbm}
\usepackage{multirow}
\usepackage{xcolor}
\usepackage{hyperref}
\usepackage{pifont}
\usepackage{xr}
\usepackage{algorithm}
\usepackage{algpseudocode}
\usepackage{longtable}

\newcommand{\stcomp}[1]{{#1}^{\mathsf{c}}}

\newcommand{\rch}[1]{\textcolor{black}{#1}}  
\newcommand{\och}[1]{\textcolor{black}{#1}}  

\newcommand{\rchnew}[1]{\textcolor{black}{#1}}  

\newcommand{\suppc}[1]{\textup{\textrm{\rchnew{supp}}}\rch{\left(#1\right)}}

\DeclareMathOperator*{\argmin}{arg\,min}

\usepackage{lastpage}
\jmlrheading{24}{2023}{1-\pageref{LastPage}}{1/22}{1/23}{22-0019}{Shiyu Duan, Spencer Chang, and Jos\'{e} C. Pr\'{i}ncipe}
\ShortHeadings{Efficient Learning Using Sufficient Labels}{Duan, Chang, and Pr\'{i}ncipe}
\firstpageno{1}

\begin{document}

\title{Labels, Information, and Computation: \\Efficient Learning Using Sufficient Labels}

\author{\name Shiyu Duan \email michaelshiyu3@gmail.com \\
        \name Spencer Chang \email chang.spencer@ufl.edu \\
       \name Jos\'{e} C. Pr\'{i}ncipe \email principe@cnel.ufl.edu \\
       \addr Department of Electrical and Computer Engineering\\
       University of Florida\\
       Gainesville, FL 32611, USA}

\editor{Isabelle Guyon}

\maketitle


\begin{abstract}
  In supervised learning, obtaining a large set of fully-labeled training data is expensive.
  We show that we do not always need full label information on every single training example to train a competent classifier.
  Specifically, inspired by the principle of sufficiency in statistics, we present a statistic (a summary) of the fully-labeled training set that captures almost all the relevant information for classification but at the same time is easier to obtain directly.
  We call this statistic ``sufficiently-labeled data'' and prove its sufficiency and efficiency for finding the optimal hidden representations, on which competent classifier heads can be trained using as few as a single randomly-chosen fully-labeled example per class. 
  Sufficiently-labeled data can be obtained from annotators directly without collecting the fully-labeled data first.
  And we prove that it is easier to directly obtain sufficiently-labeled data than obtaining fully-labeled data.
  Furthermore, sufficiently-labeled data is naturally more secure since it stores relative, instead of absolute, information.
  Extensive experimental results are provided to support our theory.
\end{abstract}


\section{Introduction}
Consider a \(c\)-class classification task with i.i.d. training data \(\{X_i, Y_i\}_{i=1}^n\), with each \(X_i\) being an input example and \(Y_i\in\{1, \ldots, c\}\) its label.
Do we always need full label information on every single training example to estimate a competent classifier on its distribution?
Despite that such practice is pervasive in machine learning, we think that more efficient strategies exist.

Our inspiration originates from the principle of sufficiency in statistics.   
On a high level, a sufficient statistic is a function of data that contains all its information when it comes to estimating an unknown parameter of the underlying distribution \citep{casella2002statistical}.
Effective data reduction can be achieved with a sufficient statistic at hand.
For example, for normal distribution, sample mean, a one-number summary, is just as informative as the whole sample for estimating population mean.
And similarly, sample variance suffices for estimating population variance.
For uniform distributions, sample minimum and maximum together carries an equal amount of information as the whole sample for estimating the support. 

In this work, we extend this principle of sufficiency to classification with a focus on reducing the cost of obtaining labeled training data.
We discover a statistic on fully-labeled data that (1) can be obtained directly from annotators without having to collect fully-labeled data first and (2) is easier to collect compared to fully-labeled data but at the same time (3) captures relevant information from data for learning the optimal hidden representations. 
Then a classifier head can be trained on the learned hidden representations using a set of fully-labeled data.
Competent classifier heads can be trained using as few as a single randomly-chosen fully-labeled example per class.
This allows one to learn performant models with less costly training data by using mostly this statistic in place of fully-labeled data.
We call this statistic ``sufficiently-labeled data'' to emphasize the connection.
And the sufficiency of this statistic can be rigorously proven in certain settings, allowing us to interpret training with sufficiently-labeled data as training using only relevant information, no more and no less.

If given a fully-labeled dataset, encoding it into a sufficiently-labeled one can be viewed as a generic encryption scheme that protects data security and user privacy. 
Specifically, a sufficiently-labeled training set contains only ``relative labels'' on user pairs.
And no absolute information on individuals can be re-identified based on these relative labels.
This is ideal in settings where the labels to be predicted on individuals contain sensitive information such as the diagnosis of a certain disease. 
In this case, sensitive user information no longer needs to be stored and/or transported, enhancing the security of the pipeline without extra overhead or compromise in performance.

\noindent\textbf{Contributions} \newline
This work makes the following theoretical contributions.
We define sufficiently-labeled data to be input example pairs with each pair having a binary label stating whether the two examples are from the same class.
We then present a framework to learn with a mixture of sufficiently-labeled and fully-labeled data.
Specifically, the hidden layers are trained using sufficiently-labeled data, and the linear output layer is then trained using fully-labeled data without fine-tuning the hidden layers.\footnote{Many popular classifier networks such as the ResNets \citep{he2016deep} have a linear output layer, so our method is widely applicable.}
We prove the following results. 
\begin{enumerate}
   \item Sufficiently-labeled data is sufficient for finding the optimal parameters of the hidden layers, and the proposed learning method produces solutions that are as competent as those produced using fully-labeled data.
   \item Having more sufficiently-labeled data reduces the need for fully-labeled data.
   \item Sufficiently-labeled data can be derived from fully-labeled data, but it can also be directly collected.
   And it is easier to collect sufficiently-labeled data than fully-labeled data. 
\end{enumerate} 

In addition to our theoretical contributions, we present the following experimental results. 
On MNIST, Fashion-MNIST \citep{xiao2017fashion}, SVHN \citep{netzer2011reading}, CIFAR-10, \rch{and CIFAR-100} \citep{krizhevsky2009learning}, we empirically verify that, given a large set of sufficiently-labeled data for training the network body, only a single randomly-selected fully-labeled example per class is needed to train a state-of-the-art classifier.
Further, our proposed learning method shares similar sample complexity as learning using purely fully-labeled data.

\noindent\textbf{Paper Structure}\newline
In Section~\ref{theory} and \ref{general algorithm}, we present our theory of sufficient labels.
How sufficient labels can be used to help protect user privacy is described in Section~\ref{privacy}.
Related work is then discussed in Section~\ref{related}.
Section~\ref{experiments} contains our experimental results.
And finally, conclusions are given in Section~\ref{conclusions}.

\section{A Theoretical Framework for Learning From Sufficient Labels}
\label{theory}

In this chapter, we present a framework for learning with a mixture of sufficiently-labeled and fully-labeled data.
The settings and notations are established first in Section~\ref{sec1}.
A learning method is presented in Section~\ref{sec6}.
We give an overview on the theoretical guarantees we provide for this method in Section~\ref{sec2}.
Then we proceed to formally present and prove these results in Section~\ref{sec3}, \ref{sec4}, and \ref{sec5}.

\subsection{Notations and Settings}
\label{sec1}
In this section, some notations are established and the settings in which our theoretical results work are described.
We also formally define our sufficient labels.
Note that the settings below are only for the theoretical results in this section.
We later generalize our learning algorithm to the general classification setup.
\rch{In particular, we first present and analyze our learning algorithm in the binary classification setting.
The algorithm is later generalized to multi-class classification and we prove its efficacy through extensive experiments, but rigorously extending the theory to the multi-class case is beyond the scope of this paper.
With that said, a theoretical analysis on the multi-class case should still be strongly related to our treatment on the binary case in this paper.
This is because the multi-class extension of our algorithm essentially treats the multi-class classification problem as a set of binary classification subproblems using the classic one-versus-the-rest strategy, and solves it using a group of one-versus-the-rest subclassifiers (this aligns with the standard practice in designing deep neural network classifiers, where each output layer node is responsible for predicting one particular class).
}

\subsubsection{Some Notations}
We use  \(\mathbbm{1}\) to denote the indicator function.
The support of a function or random element \(f\) is denoted \(\suppc{f}\).
The cardinality of a set \(\mathbb{U}\) is written as \(|\mathbb{U}|\).
For data, we use upper case letters for random elements and lower case letters for realizations of random elements.
Throughout the text, the term ``with probability'' will be abbreviated to ``w.p.''.

\subsubsection{Data}
We consider two-class classification with data being random elements \(X\in\mathbb{X}\subseteq\mathbb{R}^d, Y\in\{+, -\}\), where \(X\) is the input and \(Y\) its label, and \(\mathbb{X}\) is a subspace.
Let the density function of \((X, Y)\) be \(p_{X, Y}(x, y)\).
For any two independent \((X_1, Y_1), (X_2, Y_2)\), it is evident that 
\begin{equation}
   p_{X_1, X_2|Y_1, Y_2}(x_1, x_2) = p_{X_1|Y}(x_1) p_{X_2|Y}(x_2).
\end{equation}
Let \(\alpha=\Pr(Y=+)\in(0, 1)\).

\subsubsection{Sufficient Labels}

\begin{definition}[Sufficient Label]
   Given random variables \(Y, Y^\prime\), a sufficient label is defined as 
   \begin{equation}
      T(Y, Y^\prime) = \mathbbm{1}(Y = Y^\prime).
   \end{equation}
\end{definition}
\begin{definition}[Sufficiently-Labeled Data]
   A sufficiently-labeled dataset of size \(n\in\mathbb{N}\) is given as 
   \begin{equation}
      \left\{\left(X_{1_i}, X_{2_i}, T\left(Y_{1_i}, Y_{2_i}\right)\right)\right\}_{i=1}^n,
   \end{equation} 
   where \(\left\{\left(X_{j_i}, Y_{j_i}\right)\right\}_{j=1,2, i=1,\ldots, n}\) is a set of i.i.d. random elements sharing distribution with \((X, Y)\). 
\end{definition}

Sufficiently-labeled data can be derived from fully-labeled data.
On a fully-labeled sample of size two \(\left(X_1, Y_1\right), \left(X_2, Y_2\right)\), a sufficient label is a summary of the labels.
Specifically, it summarizes whether these two examples are from the same class.
In general, if we have a fully-labeled sample \(\left\{\left(X_i, Y_i\right)\right\}_{i=1}^{2n}\), we can reduce it into a sufficiently-labeled sample that is a summary of the original sample:
\begin{equation}
   \left\{\left(X_{1_i}, X_{2_i}, T\left(Y_{1_i}, Y_{2_i}\right)\right)\right\}_{i=1}^n.
\end{equation} 
The reduction can be performed in multiple ways by arranging the indices \(1_i, 2_i\).

On the other hand, in practice, a sufficiently-labeled sample can evidently be obtained directly without having to collect the fully-labeled dataset first. 
For this, the annotators should be presented with example pairs and should be asked to label if each pair are from the same class.
The actual class of each individual example in any pair does not have to be identified.
Conceptually, this requires less effort from annotators than asking them to fully label each individual example.
Moreover, as a theoretical result, we later justify from a learning theoretical point of view why it is easier to collect sufficiently-labeled data. 

\subsubsection{Model and Risk}

We consider models of the form \(f = f_2\circ F_1\), where \(\mathbb{F}_1\ni F_1:\mathbb{X}\to\mathbb{H}\), \(\mathbb{F}_2\ni f_2:\mathbb{H}\to\mathbb{R}\). 
\rch{Our theory works with any real inner product space \(\mathbb{H}\). For typical deep learning settings, we may consider \(\mathbb{H}\) to be \(\mathbb{R}^p\) for some \(p\) for convenience with the dot product and the \(\ell^2\) norm as the inner product and the norm of choice.}
\(\mathbb{F}_1, \mathbb{F}_2\) are some hypothesis spaces.
\rch{For binary classification, the model prediction on \(X\) is obtained as \(\text{sign}(f(X))\).}
Define 
\begin{equation}
   \mathbb{F} = \left\{f| f = f_2\circ F_1, f_2\in\mathbb{F}_2, F_1\in\mathbb{F}_1\right\}.
\end{equation}

In addition, assume that \(f_2\) is a linear model in \rch{\(\mathbb{R}^p\)}: \(f_2(\cdot) = \langle w, \phi(\cdot)\rangle\), where \(\phi\) is some feature map, and that it is parameterized by \(w\).
\(\phi\) and \(w\) are assumed to satisfy
\begin{equation}
   \label{eq1}
   \|\phi(u)\| = r, \forall u, \rch{\|w\|\leq\frac{1}{r}}.
\end{equation}

Note that this model formulation is broad enough to include the popular classifier networks such as the VGG networks \citep{simonyan2014very}, ResNets \citep{he2016deep}, and DenseNets \citep{huang2017densely}.
For these models, \(\phi\) can be the nonlinearity between the network body and the final linear layer.
\rch{If needed, one may normalize the network activation vector after this nonlinearity and the weight vector of the output layer to ensure that Eq.~\ref{eq1} is satisfied.
For example, suppose we are given activation vector \(\phi'\), weight vector \(w'\), then we may set \(\phi = \phi'r / \|\phi'\|\) and \(w = w' / \left(\|w'\|r\right)\).} 

The regular hinge loss is given as: 
\begin{equation}
\ell^{0+}:\mathbb{R}\times\{+, -\}\to\mathbb{R}:(\hat{y}, y)\mapsto \max(0, 1 - y\hat{y}). 
\end{equation}
\rch{Based on the assumption on \(f\) (Eq.~\ref{eq1}), we have
\begin{equation} 
   \left|\hat{y}\right| = \left|f(x)\right| = \left|\langle w, \phi\left(F_1(x)\right)\rangle\right| \leq \|w\|\|\phi\left(F_1(x)\right)\|\leq 1.
\end{equation}
Therefore, it is equivalent to analyze this unbounded (from below) loss:}
\begin{equation}
\ell:\mathbb{R}\times\{+, -\}\to\mathbb{R}:(\hat{y}, y)\mapsto - y\hat{y}. 
\end{equation}
Define the risk to be
\begin{equation}
   R(f_2\circ F_1, X, Y) = E_{X, Y}\ell(f_2\circ F_1(X), Y).
\end{equation}
The goal is to minimize this risk.
A sample mean estimation to this risk is given as follows.
\begin{equation}
   \hat{R}\left(f_2\circ F_1, \left\{\left(x_i, y_i\right)\right\}_{i=1}^n\right) = \frac{1}{n}\sum_{i=1}^n\ell\left(f_2\circ F_1(x_i), y_i\right),
\end{equation}
where \(\left\{\left(x_i, y_i\right)\right\}_{i=1}^n\) is a realization of an i.i.d. random sample sharing the same distribution as \((X, Y)\). 

\subsection{How to Learn With Sufficient Labels}
\label{sec6}

We present an algorithm for learning with sufficiently-labeled data.
A similar learning method was proposed for modularizing deep learning workflows \citep{duan2020modularizing}, but was not analyzed from the perspective of sufficient labels.

Suppose we are given the hypothesis space \(\mathbb{F}\). 
Also, suppose we are given training data \(\left\{\left(x_{1_i}, x^\prime_{1_i}, T\left(y_{1_i}, y^\prime_{1_i}\right)\right)\right\}_{i=1}^{n_1}\), \(\left\{x_{2_i}, y_{2_i}\right\}_{i=1}^{n_2}\), with \(\left(x_{i_j}, y_{i_j}\right)\) and \(\left(x^\prime_{1_j}, y^\prime_{1_j}\right)\) being i.i.d. sharing the same distribution as \((X, Y)\) for all \(i, j\). 

Define 
\begin{align}
   \label{eq2}
   &\ell_1\left(F_1(x), F_1\left(x^\prime\right), T\left(y, y^\prime\right)\right) = (-1)^{T\left(y, y^\prime\right) + 1}\left\| \phi\circ F_1(x) - \phi\circ F_1\left(x^\prime\right) \right\|\och{^2}.
\end{align} 
And also define a sample mean estimation
\begin{align}
   &\hat{R}_1\left(F_1, \left\{\left(x_{1_i}, x_{1_i}^\prime, T\left(y_{1_i}, y_{1_i}^\prime\right)\right)\right\}_{i=1}^{n_1}\right) = \frac{1}{n_1}\sum_{i=1}^{n_1}\ell_1\left(F_1\left(x_{1_i}\right), F_1\left(x_{1_i}^\prime\right), T\left(y_{1_i}, y_{1_i}^\prime\right)\right),
\end{align}
The learning algorithm consists of the following two steps.
\begin{enumerate}[label=\roman*]
   \item Step 1 (Training the Hidden Layers): Find an \(\hat{F}_1\) in 
   \begin{equation}
   \argmin_{F_1\in\mathbb{F}_1} \hat{R}_1\left(F_1, \left\{\left(x_{1_i}, x_{1_i}^\prime, T\left(y_{1_i}, y_{1_i}^\prime\right)\right)\right\}_{i=1}^{n_1}\right).
   \end{equation}
   \item Step 2 (Training the Output Layer): Find an \(\hat{f}_2\) in  
   \begin{equation}
      \argmin_{f_2\in\mathbb{F}_2}\hat{R}\left(f_2\circ \hat{F}_1, \left\{\left(x_{2_i}, y_{2_i}\right)\right\}_{i=1}^{n_2}\right).
   \end{equation}
   \item Returns: \(\hat{f}_2\circ\hat{F}_1\).
\end{enumerate}
Note that in Step 2, \(F_1\) is kept frozen at \(\hat{F}_1\).
Also note that Step 1 only requires sufficiently-labeled data and Step 2 fully-labeled data.

Some immediate questions follow.
(1) How do we know if and how well this learning algorithm works in terms of finding
\begin{equation}
   \argmin_{f_2\circ F_1\in\mathbb{F}}R(f_2\circ F_1, X, Y)?
\end{equation}
(2) Is it easier to collect training data for this algorithm than learning with only fully-labeled data?
These questions will be rigorously answered in the following sections.
Before diving into the detailed arguments, we first formally present the claims we wish to prove for this algorithm in the next section.
This serves as an overview on our theoretical results that helps facilitate a better high-level understanding. 

\subsection{Overview of Theoretical Results}
\label{sec2}

Suppose empirical risk minimization (ERM) on \(\hat{R}\) (learning with purely fully-labeled data) can map training data \(\left\{\left(x_i, y_i\right)\right\}_{i=1}^n\) and a hypothesis space \(\mathbb{F}\) to a solution \(F\in\mathbb{F}\) that attains at most \(\gamma\) true risk.
We can make the following claims for our learning algorithm \(\rch{\mathcal{A}_\mathrm{ours}}\) in Section~\ref{sec6}. 
\begin{enumerate}
   \item \rch{(Section~\ref{sec3})} \(\rch{\mathcal{A}_\mathrm{ours}}\) can map \(\left\{\left(x_{1_i}, x_{1_i}^\prime, T\left(y_{1_i}, y_{1_i}^\prime\right)\right)\right\}_{i=1}^{n_1}\), \(\left\{\left(x_{2_i}, y_{2_i}\right)\right\}_{i=1}^{n_2}\), and \(\mathbb{F}\) to a hypothesis \(F_{ours}\in\mathbb{F}\) with at most \(\gamma\) true risk for some \(n_1, n_2\). 
   
   \item \rch{(Section~\ref{sec4})} \(n_2\) decreases with \(n_1\). In other words, having more sufficiently-labeled data reduces the need for fully-labeled data.
   \rch{This result is important because it shows that we can not only use sufficient labels \textit{with} full labels, \rchnew{but also} use sufficient labels \textit{in place of} full labels.}

   \item \rch{(Section~\ref{sec5})} Given \(X, X^\prime\), sufficient label \(T(Y, Y^\prime)\) is easier to obtain than full labels \(Y, Y^\prime\). \rch{This can be justified from several perspectives. For example, sufficient label is conceptually easier to annotate and may simply be more suitable in certain scenarios.} From a learning theoretical standpoint, the sample complexity of learning \rch{a classifier to predict} full label exceeds that for sufficient label as the number of underlying classes increase.
\end{enumerate}
For the first claim, we empirically show that \(n_1 + n_2\) is \(\mathcal{O}(n)\).
This shows that \(\rch{\mathcal{A}_\mathrm{ours}}\) is as performant in finding the risk minimizer.
The third claim is a general one on the sufficient labels, meaning that it does not depend on the proposed learning algorithm or the settings assumed in this section.
These three statements will be proven in the subsequent three sections, respectively.

On a high level, these results together yields the following conclusions.
First, performant models can be trained efficiently with our learning algorithm using a mixture of sufficiently-labeled and fully-labeled training data.
In addition, it is easier to collect labeled training data for our learning method compared to learning with only fully-labeled data. 

\subsection{Finding the Risk Minimizer With Sufficient Labels}
\label{sec3}


To justify that the proposed learning algorithm can indeed find a minimizer for the true risk \(R\) given enough training data, we can show that its Step 1 finds an \(F_1\in\mathbb{F}_1\) for which there exists an \(f_2\in\mathbb{F}_2\) such that \(f_2\circ F_1\) is a risk minimizer. 
To this end, we have the following theorem. 
The proofs of all our theoretical results are provided in the supplementary materials.

\begin{theorem}
   \label{th1}  
   Define 
   \begin{align}
         &\mathbb{D} = \argmin_{F_1\in\mathbb{F}_1}E_{X, X^\prime|Y\neq Y^\prime} \left(-\left\| \phi\circ F_1(X) - \phi\circ F_1(X^\prime) \right\|\och{^2}\right),\\
         &\mathbb{S} = \{F_1\in\mathbb{F}_1| \Pr(\phi\circ F_1(X) = \phi\circ F_1(X^\prime)| Y=Y^\prime) \\ \nonumber
         &\qquad\qquad\qquad= \Pr(\phi\circ F_1(X) \neq \phi\circ F_1(X^\prime)| Y\neq Y^\prime) = 1\},\\
         &\mathbb{F}^\star = \left\{f_2\circ F_1\Bigg| f_2\circ F_1\in\argmin_{f_2\in\mathbb{F}_2, F_1\in\mathbb{F}_1} R(f_2\circ F_1, X, Y)\right\},\\
         &\mathbb{F}_1^\star = \left\{F_1| \exists f_2, f_2\circ F_1\in\mathbb{F}^\star\right\}.
   \end{align}
   If \(\mathbb{D}\cap\mathbb{S}\neq\emptyset\) and \(|\suppc{\phi\circ F_1(X)}| > 4, \forall F_1\notin\mathbb{S}\), then  
   \begin{equation}
      \mathbb{F}_1^\star = \mathbb{D}\cap\mathbb{S}.
   \end{equation}
\end{theorem}

Now, to show that the proposed learning method finds a true risk minimizer, it suffices to show that, given enough data, Step 1 finds an element in \(\mathbb{D}\cap\mathbb{S}\).
To see this, note that 
\begin{equation}
   \label{eq14}
   \frac{1}{n_1}\sum_{i=1}^{n_1} \mathbbm{1}\left(T\left(y_{1_i}, y_{1_i}^\prime\right) = 0\right)\left(-\left\| \phi\circ F_1\left(x_{1_i}\right) - \phi\circ F_1\left(x^\prime_{1_i}\right)\right\|\och{^2}\right)
\end{equation}
can be seen as an approximation to 
\begin{equation}
   \label{eq15}
   E_{X, X^\prime|Y\neq Y^\prime}\left(-\left\|\phi\circ F_1(X) - \phi\circ F_1\left(X^\prime\right)\right\|\och{^2}\right),
\end{equation}
and 
\begin{equation}
   \label{eq16}
   \frac{1}{n_1}\sum_{i=1}^{n_1} \mathbbm{1}\left(T\left(y_{1_i}, y_{1_i}^\prime\right) = 1\right)\left\| \phi\circ F_1\left(x_{1_i}\right) - \phi\circ F_1\left(x^\prime_{1_i}\right)\right\|\och{^2}
\end{equation}
to
\begin{equation}
   \label{eq5}
   E_{X, X^\prime|Y= Y^\prime}\left\|\phi\circ F_1(X) - \phi\circ F_1\left(X^\prime\right)\right\|\och{^2},
\end{equation}
which is minimized if and only if \(\phi\circ F_1(X) = \phi\circ F_1\left(X^\prime\right)\) w.p. \(1\) given \(Y = Y^\prime\). 
\rch{More formally, since the data pairs are assumed to be i.i.d., the law of large numbers directly applies here and states that the empirical risks (Eq.~\ref{eq14} and Eq.~\ref{eq16}) eventually converge to their expectation counterparts (Eq.~\ref{eq15} and Eq.~\ref{eq5}) as \(n_1\) grows.
More analysis on finite sample behavior can be done here, and other forms of empirical risks that enjoy better convergence properties may exist.
Further study on these topics is out of the scope of this paper and we leave it as future work.}

\subsubsection{\rchnew{Connecting Sufficiently-Labeled Data With Sufficient Statistic}}

\rchnew{In Theorem~\ref{th1}, \(\mathbb{F}_1^\star\) is reformulated with expressions involving only sufficiently-labeled data \(\left(X, X^\prime, T(Y, Y^\prime)\right)\). This implies that \(\left(X, X^\prime, T(Y, Y^\prime)\right)\) contains as much information as \((X, Y)\) for finding \(\mathbb{F}_1^\star\), which is analogous to how a sufficient statistic supplies all the information for estimating an unknown parameter \citep{casella2002statistical}.}

\rchnew{This similarity between sufficiently-labeled data and a sufficient statistic is summarized as follows.}

\rchnew{\(\bullet\) \textbf{For estimating an unknown parameter of a statistical model, a sufficient statistic is as informative as the entire sample \citep{casella2002statistical}.}}

\rchnew{\(\bullet\) \textbf{For finding the optimal classifier body, sufficiently-labeled data is as informative as fully-labeled data.}}

\subsubsection{Interpretations: Rethinking Classification in Terms of Pairwise Statistic}
\begin{figure}[t]
   \centering
   \includegraphics[width=.8\columnwidth]{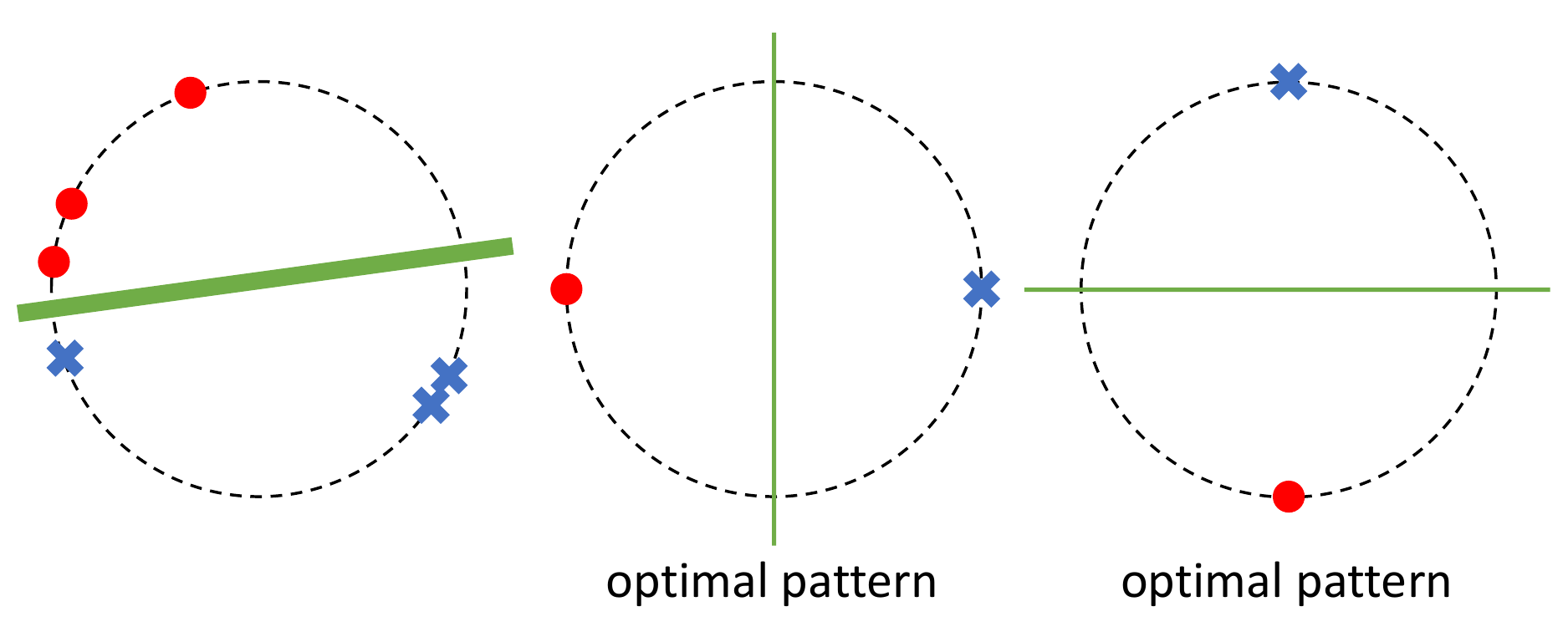}
   \caption{
      The sufficiency of sufficiently-labeled data.
      The hidden module \(F_1\) learns to generate a pattern in a feature space (illustrated here) such that the output layer \(f_2\), a linear model in this feature space, can most effectively classify.
      Due to the assumption on the feature map, i.e., \(\|\phi(u)\|= r,\forall u\), every example must be on a circle in this illustration.
      For each pattern, the \(f_2\) that achieves zero hinge loss (perfect separation) with minimum \(\|w\|\) (model capacity) is the green line.
      The weight of the line is proportional to \(\|w\|\). 
      The optimal pattern, i.e., one that allows perfect separation with the smallest possible capacity (and therefore best generalization \citep{vapnik2013nature, shalev2014understanding}), is the one where each pair of examples from the same class are mapped to the same point, whereas each pair from different classes are as far away as possible.
      This optimality is described only using pairwise relationships and sufficient labels.
      }
   \label{fig1}
\end{figure}

In classification, data examples are usually considered individually and used with full label information. 
However, our results state that pairwise statistics of data contain all relevant information for learning the optimal hidden representations, allowing for data reduction.
Intuitively, this is easy to understand.
Consider \(F_1\) to be learning a pattern in a feature space such that \(f_2\), a linear classifier in that feature space, can most effectively classify (in terms of minimizing the hinge loss).
Then as illustrated in Fig.~\ref{fig1}, how \(F_1\) arranges each individual example does not matter --- all patterns are equally optimal if (1) each pair from distinct classes are mapped as far away from each other as possible and (2) each pair from the same class are mapped to the same point.
This pattern is fully described using pairwise summary and that an optimal \(F_1\) is one that learns such a pattern is essentially what Theorem~\ref{th1} states.
\rch{Indeed, the set \(\mathbb{S}\) characterizes the set of \(F_1\) that, with probability \(1\), maps every example from the same class to the same vector representation and those from distinct classes to different ones.
On the other hand, \(\mathbb{D}\) requires that \(F_1\) maximizes (in expectation) the distance between example pairs from different classes.}

This sufficiency of \(\left(X, X^\prime, T\left(Y, Y^\prime\right)\right)\) allows us to understand it as a way to partition the sample space and leave only relevant information \citep{casella2002statistical}. 
Thus, learning with sufficiently-labeled examples directly is learning using relevant data, no more and no less.
And going from fully-labeled data to sufficiently-labeled can be interpreted as extracting relevant information prior to any training.

Finally, from the perspective of learning representations, \(\phi\circ F_1\) determines the learned internal representations of a network. 
Hence, our results show that internal representations for classification can be learned using only sufficiently-labeled data. 
Fully labeled data is only needed to find the optimal linear mapping into the label space.

\subsubsection{Sample Complexity}
Theorem~\ref{th1} guarantees that the proposed learning algorithm finds a risk minimizer given enough sufficiently-labeled and fully-labeled training data, but it does not say anything about the sample complexity of the algorithm.
We will later show via experiments that this algorithm in fact has similar sample complexity as one that leverages purely fully-labeled data.
In other words, if ERM can use \(n\) fully-labeled examples to find a solution with a certain test performance, the proposed algorithm can find a solution with similar test performance using a mixture of sufficiently-labeled and fully-labeled data, with total sample size being comparable to \(n\).

\subsection{Having More Sufficiently-Labeled Data Means Needing Fewer Fully-Labeled Data}
\label{sec4}

The proposed learning algorithm requires two sets of training data, one sufficiently-labeled, the other fully-labeled, in this order.
In this section, we show that, to attain a particular test performance, the needed number of fully-labeled training examples decreases with the number of sufficiently-labeled examples used. 
To this end, we have the following theorem.

\begin{theorem}
   \label{th2}
   Define \(\mathbb{F}_{2, A} = \{f_2|\|w\|\leq A\}\), and order these hypothesis spaces with \(A\).
   Let \(\mathbb{S}, \mathbb{D}\) be defined as in Theorem~\ref{th1} with the assumptions therein satisfied.
   And let a true risk value \(\gamma< 0\) be given.\footnote{Any positive risk value can be improved to \(0\) by the trivial solution \(w = 0\). Therefore, only negative risk values are relevant.}
   For each \(F_1\in\mathbb{F}_1\) such that there exists \(f_2\in\mathbb{F}_2\) with \(f_2\circ F_1\) attaining this true risk value, denote the smallest \(\mathbb{F}_{2, A}\) such that \(\min_{f_2\in\mathbb{F}_{2, A}}R(f_2\circ F_1, X, Y) = \gamma\) as \(\mathbb{F}_{2, F_1, \gamma}\). 
   \rch{For each such \(F_1\)}, let an \(n_2\in\mathbb{N}\) be given and define 
   \begin{equation}
      \hat{\mathbb{F}}_{2, F_1, \gamma}^\star = \argmin_{f_2\in\mathbb{F}_{2, F_1, \gamma}}\hat{R}\left(f_2\circ F_1, \left\{\left(x_{2_i}, y_{2_i}\right)\right\}_{i=1}^{n_2}\right).
   \end{equation}
   \rch{Let \(t(F_1)\) be the function that gives the smallest \(A\) such that for all \(f_2\in\mathbb{F}_{2, F_1, \gamma}\), we have \(\|w\|\leq A\)\rchnew{.}}
   \rchnew{F}or any given probability \(\delta\), we have, with probability at least \(1 - \delta\), that
   \begin{align}
      &\sup_{\hat{f}_2\in\rchnew{\hat{\mathbb{F}}_{2, F_1, \gamma}^\star}}R\left(\hat{f}_2\circ F_1, X, Y\right) - \gamma \leq \rch{2\frac{t\left(F_1\right)r}{\sqrt{n_2}}} + 5t\left(F_1\right)r\sqrt{\frac{2\ln\left(8/\delta\right)}{n_2}}.
   \end{align}

   And  
   \begin{equation}
      \label{eq17}
      \argmin_{F_1\in\mathbb{F}_1} t(F_1) = \mathbb{S}\cap\mathbb{D}.
   \end{equation}
\end{theorem}

\rchnew{To understand this theorem, first note that the norm upper bound \(A\) returned by \(t(F_1)\) cannot be arbitrarily small. 
To see this, note that for an $F_1$, $\mathbb{F}_{2, F_1, \gamma}$ is defined to contain an $f_2$ such that $f_2\circ F_1$ attains true risk value $\gamma$. 
This implicitly requires that the $\|w\|$ of this $f_2$ cannot be arbitrarily small. 
Indeed, note that the true risk of this $f_2\circ F_1$ is $-E_{X, Y} Y f_2\circ F_1(X)$ and this term can be lower bounded by $\|w\|$ multiplied by some constant using Cauchy-Schwarz inequality. 
So, if $\|w\|$ is too small, the risk value may never reach the specified $\gamma$.}

This theorem describes how the training of \(F_1\) affects the data requirement for training \(f_2\).
In particular, a small \(t(F_1)\) results in a small \(n_2\) for the right hand side of the inequality to stay fixed at a specific value.
This indicates that Step 2 needs the fewest fully-labeled data to produce a solution that attains a particular test performance if Step 1 produced a minimizer for \(t(F_1)\). 
This theorem then confirms that Step 1 indeed produces a minimizer for \(t(F_1)\) since it finds an element in \(\mathbb{S}\cap\mathbb{D}\) given enough sufficiently-labeled data.
Therefore, the more sufficiently-labeled data Step 1 can leverage, the better it can find an element in \(\mathbb{S}\cap\mathbb{D}\), and, consequently, the fewer fully-labeled training data Step 2 will need to attain a particular test performance. 

\rch{The intuition behind why Step 1 can help learn a minimizer of $t(F_1)$ can be explained as follows. Given a data representation, i.e., an $F_1$, $t(F_1)$ characterizes the model capacity needed from $f_2$ to minimize the classification risk to a certain degree. And one can solve the problem with the simplest model (smallest $t(F_1)$) when and only when given the easiest (most separable) data representation. Step 1 essentially pursues an $F_1$ that offers the most separable data representation.}

Note that this result only goes in one direction.
Specifically, the training of \(f_2\) or the amount of fully-labeled examples given does not necessarily affect the data requirement of training \(F_1\).
This is intuitive since the learning algorithm trains \(F_1\) and \(f_2\) sequentially and freezes \(F_1\) completely after its training.

\subsection{Sufficient Labels Are Easier to Obtain Directly Than Full Labels}
\label{sec5}
\rch{Sufficiently-labeled data is less costly to obtain than fully-labeled data in many ways. 
Determining whether a given pair of examples are from the same class is conceptually simpler for an annotator compared to determining the specific class of each of the examples in the pair.
For example, to tell if two car images correspond to the same car is typically much easier than specifying the make and model of car contained in each individual image.
Also, for certain sensitive topics, people may be more comfortable with reporting sufficient labels rather than full labels. 
In the United States, for example, sometimes people are unwilling to disclose whether they voted or who they voted for in elections.
Therefore, instead of surveying each person individually and asking them if they voted or who they voted for in the past election, it would perhaps make the surveyees more comfortable if they are paired with their friend and are asked if they made the same or different voting decisions.
}

Moreover, we can justify that it is simpler to collect sufficiently-labeled data from a learning theoretical perspective.
Specifically, we show that learning to produce full labels for unlabeled examples has a larger sample complexity than learning to produce sufficient labels for unlabeled example pairs as the number of underlying classes increases.
In other words, training a competent model to assign sufficient labels for example pairs would require fewer labeled training examples than training a competent model assigning full labels.

To see this, the key observation is that the Gaussian complexity of a \(c\)-class classification problem using common loss functions such as the hinge loss or cross-entropy loss is \(\mathcal{O}\left(c^2/\sqrt{n}\right)\) in general, where \(n\) is the labeled training sample size \citep{mohri2018foundations}.
Generating sufficient labels is always a two-class classification problem regardless of the number of actual classes. 
Therefore, its sample complexity is \(\mathcal{O}(1/\sqrt{n})\), whereas the sample complexity for generating full labels is \(\mathcal{O}\left(c^2/\sqrt{n}\right)\).\footnote{There is an implicit price to pay when learning to generate sufficient labels, which is that the input dimension is doubled. This increases certain terms in typical Gaussian/Rademacher complexity bounds by, e.g., addition or multiplication by a fixed scalar \citep{bartlett2002rademacher,sun2016depth}. But the increase is constant, which will be eventually dominated by the quadratic growth in \(c\).}
This justifies our claim.

\rch{Note that this result evidently only applies in the generic multi-class classification setting.
Despite the lack of rigorous theoretical justification on our algorithm in the multi-class case, our empirical results demonstrate strong performance from our method in this setting, making our method highly relevant in practice for multi-class classification.
}

\section{A More General Learning Algorithm}
\label{general algorithm}

In Section~\ref{sec6}, we proposed a learning method for learning with sufficient labels. 
The method was proposed for binary classification with the hinge loss.
And it used a specific loss function formulation in its Step 1.

In this section, we extend this algorithm to classification with \(c\) classes (\(c\geq 2\)) using arbitrary loss function. 
And we present other alternatives that can be used as the loss function in Step 1.

Here, we assume that the model admits the form
\begin{align}
   &F = F_2\circ F_1, F_1:\mathbb{X}\to\mathbb{R}^p, F_2:\mathbb{R}^p\to\mathbb{R}^c\\
   &F_2: u\mapsto \left(\left\langle w_1, \phi(u)\right\rangle + b_1, \ldots, \left\langle w_c, \phi(u)\right\rangle + b_c\right)^\top, 
\end{align}
where \(F_2\) is a vector of binary classifiers, each classifying one class versus the rest.  
The \(i\)th coordinate of \(F_2\) corresponds to the \(i\)th class.
At test time, the model outputs the class corresponding to the maximum coordinate of \(F_2\).
Clearly, many popular classifier networks including the ResNets \citep{he2016deep} admit this representation.

Let two loss functions \(\ell: \mathbb{R}^c\times\{1, \ldots, c\}\to\mathbb{R}\), \(\ell_1: \mathbb{R}^p\times\mathbb{R}^p\times\{0, 1\}\to\mathbb{R}\) be given.
For example, \(\ell\) can be softmax followed by cross-entropy.
Also suppose we have a hypothesis space and two sets of labeled data as in the case of Section~\ref{sec6} except that now there are potentially more than two classes in the dataset. 
The two steps of this more general algorithm are as follows.

\begin{enumerate}[label=\roman*]
   \item Step 1 (Training the Hidden Layers): Find an \(\hat{F}_1\) in 
   \begin{equation}
      \label{eq4}
      \argmin_{F_1\in\mathbb{F}_1}\frac{1}{n_1}\sum_{i=1}^{n_1}\ell_1\left(F_1\left(x_{1_i}\right), F_1\left(x_{1_i}^\prime\right), T\left(y_{1_i}, y_{1_i}^\prime\right)\right).
   \end{equation} 
   \item Step 2 (Training the Output Layer): Find an \(\hat{F}_2\) in 
   \begin{equation}
      \argmin_{F_2\in\mathbb{F}_2}\frac{1}{n_2}\sum_{i=1}^{n_2}\ell\left(F_2\circ\hat{F}_1\left(x_{2_i}\right), y_{2_i}\right).
   \end{equation} 
   \item Return: \(\hat{F}_2\circ\hat{F}_1\).
\end{enumerate}

Clearly, when \(c=2\), implementing \(F_2\) as a real-valued function without bias, choosing \(\ell_1\) as in Eq.~\ref{eq2}, and then letting \(\ell\) be the unbounded binary hinge loss in Section~\ref{sec1} recovers the algorithm proposed in Section~\ref{sec6}.   

In addition to the \(\ell_1\) we used in Section~\ref{sec6}, there are other alternatives that can learn an \(F_1\) that satisfies the requirements in Theorem~\ref{th1} when plugged into Eq.~\ref{eq4}. 
Some of these alternatives are based on the ``proxy objectives'' proposed by \citet{duan2020modularizing}.

Define bivariate function \(k(u, v) = \langle \phi(u), \phi(v) \rangle\).
Let \(\beta := \rch{\min_{u, v} k(u, v)}\) and define \(k_i^\star = \mathbbm{1}\left(T\left(y_{1_i}, y_{1_i}^\prime\right) = 1\right) r^2 + \mathbbm{1}\left(T\left(y_{1_i}, y_{1_i}^\prime\right) = 0\right) \beta\).
Denote the vector in \(\mathbb{R}^{n_1}\) whose \(i\)th element is \(k\left(F_1\left(x_{1_i}\right), F_1\left(x_{1_i}^\prime\right)\right)\) as \(K_{F_1}\) and the vector whose \(i\)th element is \(k_i^\star\) as \(K^\star\).

\begin{itemize}
    \item Negative cosine similarity (NCS)\footnote{When plugged into Eq.~\ref{eq4}, the empirical risk to be minimized becomes the negated cosine similarity between \(K_{F_1}\) and \(K^\star\), hence the name. In the kernel learning literature, a loss similar to this is known as kernel alignment \citep{cristianini2006kernel}. In the information theoretical learning literature, another similar loss is known as Cauchy-Schwarz divergence \citep{principe2000information}.}: 
    \begin{align}
      &\ell_1\left(F_1\left(x_{1_i}\right), F_1\left(x_{1_i}^\prime\right), T\left(y_{1_i}, y_{1_i}^\prime\right)\right) = -n_1\frac{k\left(F_1\left(x_{1_i}\right), F_1\left(x_{1_i}^\prime\right)\right)k_i^\star}{\left\|K_{F_1}\right\|_2\left\|K^\star\right\|_2}
    \end{align}
    \item Contrastive\footnote{This is an extension of a popular loss function from contrastive learning \citep{arora2019theoretical}.}:
    \begin{align}
      &\ell_1\left(F_1\left(x_{1_i}\right), F_1\left(x_{1_i}^\prime\right), T\left(y_{1_i}, y_{1_i}^\prime\right)\right) = -\log\left(\frac{\sum_{i=1}^{n_1} \mathbbm{1}\left(T\left(y_{1_i}, y_{1_i}^\prime\right) = 1\right) e^{k\left(F_1\left(x_{1_i}\right), F_1\left(x_{1_i}^\prime\right)\right)}}{\sum_{i=1}^{n_1} e^{k\left(F_1\left(x_{1_i}\right), F_1\left(x_{1_i}^\prime\right)\right)}}\right)
    \end{align}
    \item Mean Squared Error:
    \begin{align}
      &\ell_1\left(F_1\left(x_{1_i}\right), F_1\left(x_{1_i}^\prime\right), T\left(y_{1_i}, y_{1_i}^\prime\right)\right) = \left(k\left(F_1\left(x_{1_i}\right), F_1\left(x_{1_i}^\prime\right)\right) - k_i^\star\right)^2
    \end{align}
\end{itemize}

\rch{We do not offer rigorous theoretical justification on these extensions of our algorithm.
Nevertheless, their strong performance on challenging large-scale datasets (as we shall demonstrate later) hints that more general versions of our theoretical results exist for these more generic learning settings.
For example, our earlier theoretical analysis may be extended to this multi-class classifier model by noting that each of the output node is an one-class-versus-the-rest (binary) classifier.
We leave this as future work.
}

\section{Class-Encrypted Learning With Sufficient Labels}
\label{privacy}
In important domains including finance and healthcare, data security and user privacy protection need to (and are sometimes required to) be taken into consideration when developing machine learning solutions \citep{dwork2014algorithmic, kaissis2020secure}.
In particular, traditional training datasets for supervised learning are especially vulnerable to label information leaks because they associate each individual example directly with its full label, giving away all information the moment the database is breached.
Techniques such as federated learning \citep{konevcny2015federated,konevcny2016federated}, differential privacy \citep{dwork2014algorithmic}, and homomorphic encryption \citep{acar2018survey} have been proposed to address various different aspects of security and privacy.
In this section, we discuss how converting full labels into sufficient labels can serve as a generic encryption scheme that is orthogonal and complementary to the established privacy-preserving techniques.
Note that the notion of privacy preservation we discuss here is a generic one and should not be confused with the more specific one studied in differential privacy \citep{dwork2014algorithmic}.

Sufficiently-labeled data can naturally serve as a layer of encryption on user information that is impossible to fully penetrate using only this given data.
This can be easily understood through an example.

Suppose one wants to train a predictor for some disease using a database of real patient records with each labeled with a doctor's diagnosis.
Such a label is certainly sensitive information and to protect data security and user privacy, the hospital, where the diagnoses were done and the full results known, can convert the majority of fully-labeled patient data locally to sufficiently-labeled data before transporting and/or storing this data for training.
Suppose the hospital follows the process described in Algorithm \ref{alg1}, which ensures that no single individual appears twice in the sufficiently-labeled set and that the patients in the sufficiently-labeled set do not overlap with those in the fully-labeled set.
\begin{algorithm}
   \caption{An example algorithm for converting part of a fully-labeled set of patient records into a sufficiently-labeled one}\label{alg1}
   \begin{algorithmic}
      \Require Size of sufficiently-labeled training set \(n\geq 1\), fully-labeled patient records \(P:=\{\left(x_i, y_i\right)\}_{i=1}^N, N>2n,\) with each \(x_i\) corresponding to a different individual and \(y_i\) his/her diagnosis
      \State \(a\gets 0\)
      \State \(Q\gets\emptyset\)
      \While{\(a<n\)}
         \State Randomly choose \(\left(x_b, y_b\right)\in P \; s.t.\; x_b\neq x_a\)
         \State \(Q\gets Q\cup \{\left(x_a, x_b, T\left(y_a, y_b\right)\right)\}\)
         \State \(P\gets P\setminus \{\left(x_a, y_a\right), \left(x_b, y_b\right)\}\)
         \State \(a\gets a+1\)
      \EndWhile
      \State \Return A sufficiently-labeled set \(Q\)
   \end{algorithmic}
\end{algorithm}
Then, class information of users in the sufficiently-labeled set is encrypted.
Specifically, for any individual in the sufficiently-labeled set, people with access to the data can at most know if this individual has identical result with another random individual, but not the actual diagnosis itself.

\begin{figure}[t]
   \centering
   \includegraphics[width=.8\columnwidth]{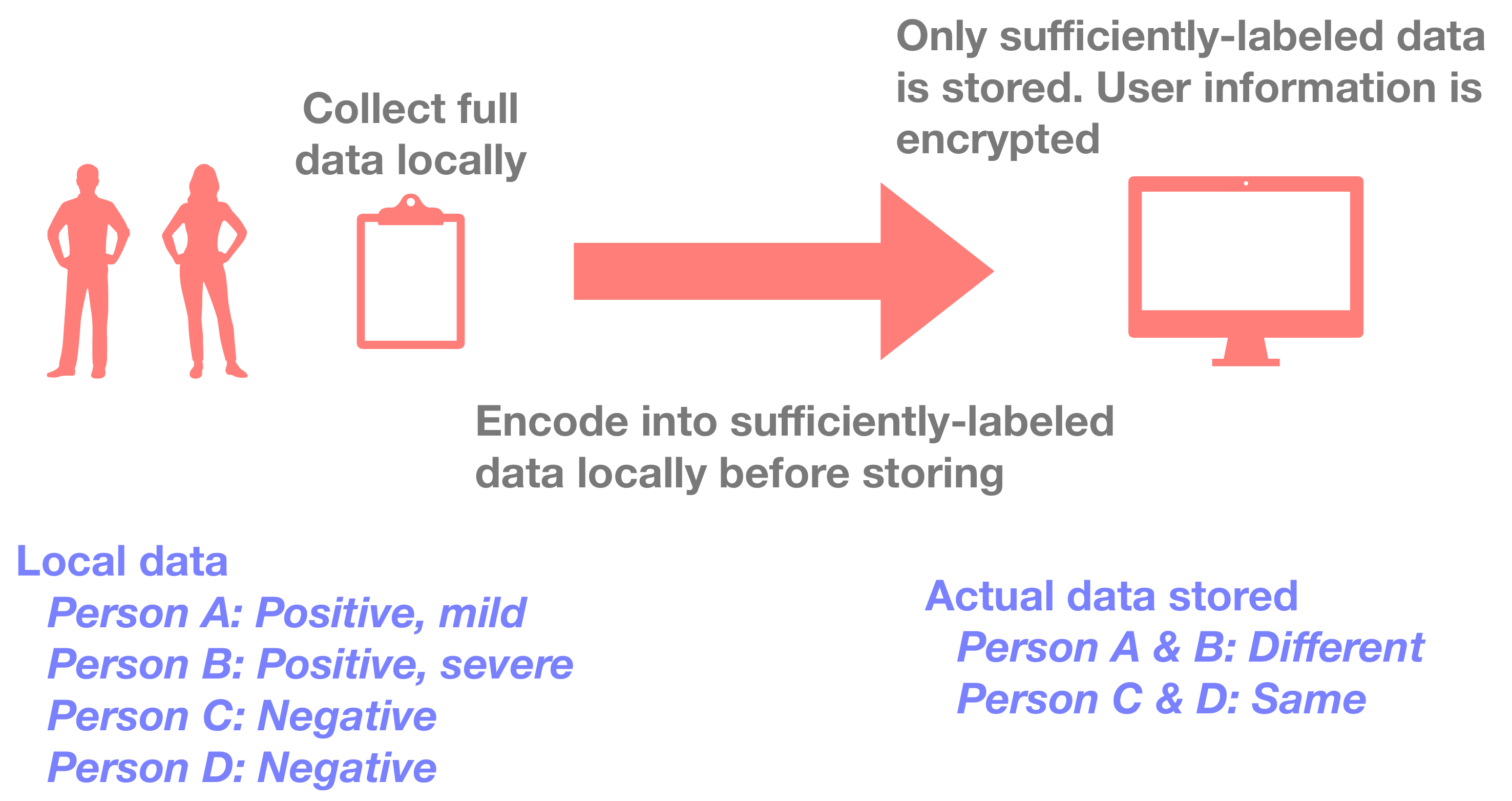}
   \caption{
      A sufficiently-labeled dataset can serve as a natural layer of encryption on sensitive user information that cannot be fully cracked. 
      In this disease diagnosis example, even if a malicious user gains full access to the sufficiently-labeled dataset, this person can at most see if two randomly-paired individuals have the same result, but not the actual diagnoses themselves.
      Such a dataset can still be used to train a performant predictor for diagnosing this disease.
      }
   \label{fig2}
\end{figure}

As we have shown, this sufficiently-labeled data (the \(Q\) returned by Algorithm \ref{alg1}), together with a small subset of unconverted fully-labeled data (the rest of \(P\) after performing Algorithm \ref{alg1}), suffices to train performant predictors.
The fully-labeled set can be as small as one example per class.
This pipeline is illustrated in Fig.~\ref{fig2}.

In general, in a sufficiently-labeled dataset \(\left\{\left(x_i, x_i^\prime, T\left(y_i, y_i^\prime\right)\right)\right\}_{i=1}^n\), the actual values of \(y_i\) and \(y_i^\prime\) cannot be re-identified if only given \(T\left(y_i, y_i^\prime\right)\).
This can help protect privacy if \(y_i\) represents some sensitive information about user \(x_i\).
In this case, only relative information between user pairs is stored in the dataset, and absolution class information of each individual user is effectively encrypted and cannot be recovered even when the dataset is breached.
This layer of encryption requires very little extra computational resource, and according to our theory, one can train models on this encrypted data without compromising performance.\footnote{Our learning method still requires a small set of fully-labeled data that would reveal \(y_i\) for each user in this set. But we have shown that this set can be as small as a single randomly-chosen example per class. Protecting the safety of such a small dataset is a much easier task.}
And it works regardless of the number of classes, contrasting some other forms of privacy-preserving labels that cease to be effective in certain cases such as binary classification \citep{ishida2017learning}.
Obviously, sufficient labels can be used alongside many existing privacy-protecting techniques for enhanced security. 

\subsection{Encryption Strength}
Given some fully-labeled user data \(P:=\{\left(x_i, y_i\right)\}_{i=1}^N, x_i\neq x_j, \forall i\neq j\), there are many ways of encoding them into a sufficiently-labeled dataset \(Q\), \footnote{We assume here that each user \(x_a\) from \(P\) appears at least once in \(Q\) (so that we are not wasting data) and that for any \(\left(x_a, x_b, T\left(y_a, y_b\right)\right)\) in \(Q\), we have \(a\neq b\).} and not all of them have the same encryption strength.

At one extreme, each user \(x_a\) in \(P\) appears \(N-1\) times in \(Q\).
In other words, we form \(Q\) from pairwise combinations of all distinct examples in \(P\).
This will give the largest \(Q\) (it contains \(N\times (N-1)/2\) examples).
But the encryption strength is the weakest in the sense that, because pairwise relationships \rch{among} all examples are available in \(Q\), one can group examples into clusters with each cluster containing all examples from one class and one class only.
Nevertheless, the actual class label of each cluster remains unknown.
But if some prior knowledge on the class distribution is given, one may succeed in guessing the true label of each cluster.

At the other extreme, each \(x_a\) in \(P\) appears exactly once in \(Q\) (one process that will lead to this case is Algorithm \ref{alg1}).
In this case, the converted \(Q\) will be the smallest (it contains \(\lfloor N/2\rfloor\) examples).
On the other hand, this provides the strongest encryption among all possible \(Q\)s if each \(x_a\) is paired with a randomly-chosen \(x_b\).
Specifically, all that one can deduce about any \(x_a\) using this dataset is whether it is from the same class as some other random \(x_b\).
And evidently, the more balanced the class distribution is, the less revealing this piece of information about \(x_a\) (and \(x_b\)) will be.

We leave as future work a rigorous study on the quantitative encryption strength of our method and its connections with other established privacy-preserving techniques.

\section{Related Work}
\label{related}

\subsection{Learning With Easier-to-Obtain Training Data}
In classification, designing learning algorithms that can produce performant models using training data that is easier to obtain has been a long-sought goal.
Depending on how ``easier to obtain'' is interpreted, we can classify previous works into the following two categories.
Works such as that by \citet{sohn2020fixmatch} do not change the underlying learning algorithms, which rely on fully-labeled data, but instead seek to reduce the cost of obtaining labels by synthesizing them instead.
On the other hand, \citet{ishida2017learning,arora2019theoretical} and others propose novel learning algorithms that can leverage information from other forms of labeled data to reduce need for fully-labeled data.
These alternative forms of labeled data are easier to obtain directly.
We review these two categories of methods separately.

\subsubsection{Reducing the Cost of Obtaining Full Labels}
A popular way to reduce the amount of human supervision needed is to synthesize (full) labels on a large set of unlabeled data by leveraging a smaller fully-labeled set and some prior knowledge.
The synthesized full labels are then used for training \citep{sohn2020fixmatch,lee2013pseudo,sajjadi2016regularization,laine2016temporal,berthelot2019mixmatch,berthelot2019remixmatch,miyato2018virtual,xie2019unsupervised,tarvainen2017mean,sajjadi2016mutual,xie2020self}.
Some of the popular approaches to obtain these synthesized labels including using network's predictions on unlabeled data \citep{lee2013pseudo,laine2016temporal}, using network's internal randomness such as dropout to implicitly augment labeled data \citep{sajjadi2016regularization}, and actively augmenting examples within classes \citep{sohn2020fixmatch,xie2019unsupervised,berthelot2019remixmatch,xie2019unsupervised,miyato2018virtual}.
State-of-the-art results are often obtained by combining multiple methods for synthesizing labels \citep{berthelot2019mixmatch,sohn2020fixmatch,xie2019unsupervised,berthelot2019remixmatch}.
These methods are sometimes called ``semi-supervised'' learning \citep{oliver2018realistic}.

These methods can be seen as ones that reduce the cost for obtaining fully-labeled training data, and are therefore orthogonal to our approach, which aims at learning from a novel form of labeled data (sufficiently-labeled) that is intrinsically easier to obtain than fully-labeled data.
Despite the same goal of facilitating learning with easier-to-obtain training data, these methods ultimately rely on the assumption that the unlabeled training set is strongly related to the fully-labeled one, which is fundamentally task-dependent.
In fact, this assumption can be easily violated in real-world datasets, in which case the synthesized labels are so noisy that they poison training rather than helping it \citep{oliver2018realistic}. 
In contrast, by approaching from a different angle, our method is inherently immune to this issue, and task-agnostic performance bounds have been drawn.
As future work, our method may be combined with the above reviewed methods such that the cost of obtaining sufficient labels can be reduced as well.

\subsubsection{Leveraging Other Forms of Labeled Data}
\label{sec7}
These works proposed new learning algorithms that can directly learn from labeled data of other forms in addition to fully-labeled data.
These novel supervision forms include, for example, pairwise (dis)similarities, complementary labels, partial labels, and even implicit labels. 

\noindent\textbf{Learning from pairwise similarities and dissimilarities.}
Some works explored learning for classification using data pairs each with a binary label indicating whether this pair of examples are from the same class or not \citep{bao2018classification,shimada2021classification,shi2021triply,hsu2018multi,bao2020similarity,balcan2008improved,bellet2012similarity}. 
These pairs are called similar or dissimilar pairs in these works, depending on if they are from the same class.
Pairwise similarities and dissimilarities are the same as sufficient labels. 
Therefore, these works are similar to ours.

\citet{bao2018classification} derived an unbiased estimator to the true classification risk using terms only involving similar pairs and unlabeled data.
They called this learning setting similar-unlabeled (SU) classification.
\citet{shi2021triply} proposed an improved optimization scheme for SU classification.
\citet{shimada2021classification} generalized the SU setting to also include dissimilar pairs, leading to the so-called similar-dissimilar-unlabeled classification problem.
The authors proposed estimators to the true classification risk that involved only similar, dissimilar, and unlabeled data.
These estimators were shown to lead to better error bounds than that obtained from SU classification.
In the above works, only mild assumptions were made on the loss function for the theoretical results to hold.
\citet{hsu2018multi} derived an empirical estimator to a specific classification risk (the negative log-likelihood (NLL)) using terms involving only similar and dissimilar pairs.
\rch{For a specific binary classification loss function, \citet{bao2020similarity} established that the classification risk can be reformulated into a risk that only involves sufficiently-labeled data up to label flipping on the classifier.}
\rch{\citet{balcan2008improved} defined \((\epsilon, \gamma, \tau)\)-goodness, which is a characterization of representations using only sufficiently-labeled data. Then they showed that for \((\epsilon, \gamma, \tau)\)-good representations, a linear classifier that optimizes the hinge loss exists with high probability.
Below, we discuss how these prior works differ from ours.}

\rch{\citet{hsu2018multi} and \citet{bao2020similarity} both demonstrated the validity of learning with a mixture of sufficiently-labeled data and fully-labeled data, echoing our work.\footnote{\rch{\cite{hsu2018multi} required full labels for learning the mapping between output nodes and semantic classes. \cite{bao2020similarity} proposed to use class priors, which requires fully-labeled data to estimate in practice.}}
However, these works are limited in the following regards.
First, neither work discussed how the sample size of the sufficiently-labeled set affects that of the fully-labeled set.
In contrast, \rchnew{w}e proved that having more sufficiently-labeled data reduces the need for fully-labeled data. 
This is important because it shows that we can not only use sufficient labels \textit{with} full labels, \rchnew{but also} use sufficient labels \textit{in place of} full labels.
In practice, we demonstrated that, with enough sufficiently-labeled data, one only needs as few as one randomly-chosen full label per class to obtain good performance. 
Successful training with such a minimal need for full labels has not been shown in these prior works, and it further proves the efficacy of learning with sufficient labels.
Second, \citet{hsu2018multi} did not discuss the sample complexity of learning with sufficiently-labeled data versus that of learning with purely fully-labeled data.
While \citet{bao2020similarity} offered such an analysis, they did not scale their evaluations beyond toy data.
Sample complexity is a key characterization of efficient learning, and describes how ``informative'' for training one sufficiently-labeled datum is compared to a fully-labeled one.
Through extensive evaluations, we established that learning from sufficient labels enjoys similar sample complexity as learning from purely full labels even on challenging high-dimensional datasets.
These are important results that make sufficient labels even more of a compelling alternative to full labels.
We also drew connection between sufficient labels and sufficient statistics, offering a new perspective on this form of supervision.
Albeit not the main focus on the paper, we pointed out the potential value of sufficient labels as an encryption method.
}

\citet{bao2018classification,shimada2021classification} focused on learning with a combination of sufficiently-labeled and unlabeled data. 
And they compared only with unsupervised and semi-supervised baselines that learned with the same data.
In comparison, we aim at learning with sufficiently-labeled and a small set of fully-labeled data.
Further, our proposed learning scheme leads to performance similar to supervised learning using full labels --- a much stronger baseline.
We think that their assumption of learning without any full label is of theoretical value but somewhat overly restrictive in practice.
Even though it is true that obtaining a large set of fully-labeled training data can be time and resource-consuming, obtaining a small set with a few examples per class is usually a trivial task.
\citet{bao2018classification,shimada2021classification} also only considered binary classification.

\rch{\((\epsilon, \gamma, \tau)\)-goodness \citep{balcan2008improved,bellet2012similarity} and variants \citep{wang2009theory} are characterizations of the optimal representation using only sufficiently-labeled data, parallel to our characterization (Theorem~\ref{th1}).
Their optimality guarantee is also similar to ours: That an optimal (in terms of, e.g., hinge loss) downstream linear classifier exists.
\((\epsilon, \gamma, \tau)\)-goodness framework requires strong constraint on the hypothesis space of \(F_1\), i.e., that it must be formed from a bivariate function (metric) evaluated on some training examples (Theorem 1 in \citep{bellet2012similarity}).
Practical algorithms such as that proposed by \citet{bellet2012similarity} also impose parametric assumptions on the form of this bivariate function.
In comparison, our theoretical analysis is generic in these regards.}

\noindent\textbf{Learning from complementarily-labeled data.}
A complementary label on example \(X\) is a class to which \(X\) does not belong.
\citet{ishida2017learning} proposed a framework to learn from complementarily-labeled data --- training sets in which each example is given one complementary label.
Upon some assumptions on data distribution and loss function formulation, a family of empirical risks using only complementarily-labeled data were shown to be unbiased estimators to the true classification risk. 
Therefore, learning to minimize these empirical risks produces solutions that minimize the true risk, given enough data.
\citet{ishida2019complementary} extended this work to arbitrary losses. 
\citet{feng2020learning} tackled the complementary label learning setting where multiple complementary labels were available for each example. 
All above works need complementarily-labeled data only but can be modified to also utilize fully-labeled data.
\citet{yu2018learning} generalized the work by \citet{ishida2017learning} to the case where, given the true label, the distribution of the complementary labels can be arbitrary rather than uniform. 
Estimation of a transition matrix requires fully-labeled data.

Complementarily-labeled data is a function of fully-labeled data, which makes it a statistic, similar to our sufficiently-labeled data.
A complementary label, like a sufficient label, is also easier to obtain than a full label.

\noindent\textbf{Contrastive learning.}
Contrastive learning methods leverage implicitly-labeled data for pre-training the network body.
Then a network head, usually a single linear layer, is trained on top of the network body using fully-labeled data for some downstream task \citep{arora2019theoretical}.
The main benefit is that the sample complexity for learning the network head can be reduced due to contrastive pre-training.
Contrastive learning methods are characterized by the use of a contrastive loss on example pairs when training the network body, and have seen great success in natural language processing \citep{mikolov2013distributed,logeswaran2018efficient,pagliardini2017unsupervised} and computer vision \citep{wang2015unsupervised}.

The data used in contrastive pre-training is actually unlabeled for the downstream task.
But we consider them to have been implicitly labeled since each pair of examples are considered similar (in the same class) or dissimilar (not in the same class) according to some prior knowledge such as co-occurrence, which is a form of (noisy) implicit label on pairs. 
In fact, theoretical guarantees have only be made when strong connections between these implicit classes and the classes in the downstream task can be assumed \citep{arora2019theoretical}. 

Contrastive pre-training does not require explicitly-labeled examples \rch{with sufficient labels}, whereas our learning method does.
In terms of the connections between this work and contrastive learning, a typical contrastive loss enforces examples from the same (implicit) class to be mapped closer to each other in the final feature space and examples from different classes to be mapped far apart \citep{arora2019theoretical}.
And this is exactly what our learning algorithm attempts to do when training the network body, and the optimality of this objective in our setting is guaranteed by our Theorem~\ref{th1}.
Thus, our learning framework can be viewed as a supervised learning analog of contrastive learning.
Further, the fact that having more sufficient labels reduces the need for full labels when training the output layer echos the result that contrastive learning reduces the labeled sample complexity of the downstream task \citep{arora2019theoretical}.

\noindent\textbf{Learning under imperfect supervision.}
Many existing works consider situations where supervision is imperfect.
In practice, slightly imperfect supervision can be much less costly to obtain, making these methods highly relevant.
One example is when only a specific class can be annotated effectively.
Another is when the labels are noisy.

Partial label learning addresses the learning settings where multiple labels are available for each instance, but only one of them is correct \citep{cour2011learning,feng2018leveraging,feng2019partiala,feng2019partialb,zhang2015solving}.
Noisy label learning assumes that there is a certain degree of error in the labels, which is common in practice \citep{han2018co,han2018masking,menon2015learning,wei2020combating,xia2019anchor}.
Learning from positive unlabeled data is to learn from only examples from a specific class and unlabeled data in a binary classification setting \citep{du2014analysis,du2015convex,elkan2008learning,kiryo2017positive,sakai2017semi,sakai2018semi}. 
This is relevant in practice, for example, when the cost for annotating a specific class is prohibitively high.
\rch{\citet{lu2019minimal} considered binary classification and proposed a setting in which the (unlabeled) training data is sampled from two marginals with complementary class priors (requires labeled data to approximate in practice).
Specifically, the first set of \(X\) should be sampled from a marginal distribution with \(p(Y=+1)=\theta\) and the second set from a marginal with \(p(Y=+1)=\theta'\neq \theta\).
Then they showed that, if the true class prior \(p(Y=+1)\) of the test distribution is known, one can derive an unbiased risk estimator using only these two sets of training data.
This setting can be viewed as sampling two fully-labeled training sets with noisy labels, with the first set of examples from class \(Y=+1\) having \(1 - \theta\) probability of incorrect labels, and the other set of examples from class \(Y=-1\) having \(\theta'\) incorrect label probability.
In practice, the best performance is obtained with \(\theta\) being close to \(1\) and \(\theta'\) close to \(0\), i.e., when the noise level is low.
}

We address the problem of learning from a new source of supervision that we call sufficient labels.
\rch{If we understand the notion of imperfect supervision to be one that includes any information source that is weaker and less complete than full labels, then we can regard sufficient labels as a type of imperfect supervision since they are summaries (less comprehensive) of full labels.}

\subsection{Modular Training of Deep Architectures}

Our proposed learning algorithms are modular in the sense that the underlying network is viewed as two components that are trained separately.
Similar provably optimal modular learning algorithms have been proposed by, e.g., \citet{duan2019kernel, duan2020modularizing}.
These methods also used a form of labeled training data similar to our sufficiently-labeled data.
However, these works analyzed the learning algorithms from the perspective of modular training, not the training data used.
In contrast, this work gives a complete analysis on the efficacy of learning with sufficient labels.
And, unlike these works, our theoretical results are not based on the assumption that the sufficiently-labeled data for training the hidden layers are derived from the fully-labeled data for training the output layer. 
This enables optimality guarantees in settings where the sufficiently-labeled data are obtained directly from annotators without collecting the more expensive fully-labeled data first.

\section{Experiments: Training With (Almost Only) Sufficiently-Labeled Data}
\label{experiments}

\begin{table*}[htbp]
\thispagestyle{empty}
   \vspace{-0.25in}
   \centering
   \vspace{-0.43in}
   \caption{Mean accuracy \(\pm\) standard deviation from \(5\) trials.
      Best result in bold. 
      For ``Data Usage'', ``Full'' refers to the number of fully-labeled examples, and ``Suff.'' the number of sufficiently-labeled examples. 
      For the models that used \(10/100\) fully-labeled examples, these examples were randomly picked at each trial, with one example from each class.
      Strong classifiers can be learned from almost entirely sufficiently-labeled data.
      } 
   \begin{tabular}{ |c|c|c|c|c| }
      \hline
      \multirow{2}{*}{Dataset (Model Used)} & \multirow{2}{*}{Loss} & \multicolumn{2}{|c|}{Data Usage} & \multirow{2}{*}{Test Acc. (\%)} \\ \cline{3-4}
      & & Full & Suff. & \\  
      \hline
      \multirow{8}{*}{MNIST (LeNet-5)} & \multirow{4}{*}{Hinge} & \(60\rch{\text{k}}\) & \(0\) & \(99.22\pm 0.05\)\\ \cline{3-5}  
      & & \(10\) & \(120\rch{\text{k}}\) & \(98.97\pm 0.15\) \\ \cline{3-5}  
      & & \(10\) & \(240\rch{\text{k}}\) & \(99.12\pm 0.08\) \\ \cline{3-5}  
      & & \(10\) & online & \(\mathbf{99.23\pm 0.05}\) \\ \cline{2-5}  
      & \multirow{4}{*}{Cross-Entropy} & \(60\rch{\text{k}}\) & \(0\) & \(\mathbf{99.32\pm 0.08}\)\\ \cline{3-5}  
      & & \(10\) & \(120\rch{\text{k}}\) & \(98.98\pm 0.15\) \\ \cline{3-5}  
      & & \(10\) & \(240\rch{\text{k}}\) & \(99.12\pm 0.07\) \\ \cline{3-5}  
      & & \(10\) & online & \(99.23\pm 0.05\) \\  
      \hline
      \multirow{8}{*}{Fashion-MNIST (ResNet-18)} & \multirow{4}{*}{Hinge} & \(60\rch{\text{k}}\) & \(0\) & \(\mathbf{95.14\pm 0.12}\)\\ \cline{3-5}  
      & & \(10\) & \(120\rch{\text{k}}\) & \(93.85\pm 0.29\) \\ \cline{3-5}  
      & & \(10\) & \(240\rch{\text{k}}\) & \(94.61\pm 0.17\) \\ \cline{3-5}  
      & & \(10\) & online & \(95.03\pm 0.28\) \\ \cline{2-5}  
      & \multirow{4}{*}{Cross-Entropy} & \(60\rch{\text{k}}\) & \(0\) & \(\mathbf{95.10\pm 0.18}\)\\ \cline{3-5}  
      & & \(10\) & \(120\rch{\text{k}}\) & \(93.89\pm 0.32\) \\ \cline{3-5}  
      & & \(10\) & \(240\rch{\text{k}}\) & \(94.63\pm 0.15\) \\ \cline{3-5}  
      & & \(10\) & online & \(95.03\pm 0.27\) \\  
      \hline
      \multirow{8}{*}{SVHN (ResNet-18)} & \multirow{4}{*}{Hinge} & \(73,257\) & \(0\) & \(96.49\pm 0.07\)\\ \cline{3-5}  
      & & \(10\) & \(146,514\) & \(95.77\pm 0.12\) \\ \cline{3-5}  
      & & \(10\) & \(293,028\) & \(96.11\pm 0.11\) \\ \cline{3-5}  
      & & \(10\) & online & \(\mathbf{96.66\pm 0.16}\) \\ \cline{2-5}  
      & \multirow{4}{*}{Cross-Entropy} & \(73,257\) & \(0\) & \(96.46\pm 0.10\)\\ \cline{3-5}  
      & & \(10\) & \(146,514\) & \(95.78\pm 0.11\) \\ \cline{3-5}  
      & & \(10\) & \(293,028\) & \(96.13\pm 0.10\) \\ \cline{3-5}  
      & & \(10\) & online & \(\mathbf{96.67\pm 0.16}\) \\  
      \hline
      \multirow{8}{*}{CIFAR-10 (ResNet-18)} & \multirow{4}{*}{Hinge} & \(50\rch{\text{k}}\) & \(0\) & \(94.09\pm 0.21\)\\ \cline{3-5}  
      & & \(10\) & \(100\rch{\text{k}}\) & \(90.99\pm 0.33\) \\ \cline{3-5}  
      & & \(10\) & \(200\rch{\text{k}}\) & \(93.56\pm 0.24\) \\ \cline{3-5}  
      & & \(10\) & online & \(\mathbf{94.24\pm 0.25}\) \\ \cline{2-5}  
      & \multirow{4}{*}{Cross-Entropy} & \(50\rch{\text{k}}\) & \(0\) & \(94.19\pm 0.13\)\\ \cline{3-5}  
      & & \(10\) & \(100\rch{\text{k}}\) & \(91.00\pm 0.34\) \\ \cline{3-5}  
      & & \(10\) & \(200\rch{\text{k}}\) & \(93.63\pm 0.17\) \\ \cline{3-5}  
      & & \(10\) & online & \(\mathbf{94.24\pm 0.24}\) \\  
      \hline
      \multirow{8}{*}{\rch{CIFAR-100 (ResNet-18)}} & \multirow{4}{*}{\rch{Hinge}} & \(\rch{50}\rch{\text{k}}\) & \(\rch{0}\) & \(\rch{73.22\pm 0.36}\)\\ \cline{3-5}
      & & \(\rch{100}\) & \(\rch{400}\rch{\text{k}}\) & \(\rch{70.45\pm 0.17}\) \\ \cline{3-5} 
      & & \(\rch{100}\) & \(\rch{800}\rch{\text{k}}\) & \(\rch{72.67\pm 0.32}\) \\ \cline{3-5} 
      & & \(\rch{100}\) & \rch{online} & \(\mathbf{\rch{73.79\pm 0.46}}\) \\ \cline{2-5} 
      & \multirow{4}{*}{\rch{Cross-Entropy}} & \(\rch{50}\rch{\text{k}}\) & \(\rch{0}\) & \(\mathbf{\rch{74.18\pm 0.15}}\)\\ \cline{3-5} 
      & & \(\rch{100}\) & \(\rch{400}\rch{\text{k}}\) & \(\rch{70.58\pm 0.24}\) \\ \cline{3-5} 
      & & \(\rch{100}\) & \(\rch{800}\rch{\text{k}}\) & \(\rch{72.67\pm 0.28}\) \\ \cline{3-5} 
      & & \(\rch{100}\) & \rch{online} & \(\rch{73.98\pm 0.50}\) \\ 
      \hline
   \end{tabular}
   \label{table1}
\end{table*}

In this section, we verify our first two claims in Section~\ref{sec2} through experiments. 
Specifically, we test for both the hinge loss and the cross-entropy loss on MNIST, Fashion-MNIST \citep{xiao2017fashion}, SVHN \citep{netzer2011reading}, CIFAR-10, \rch{and CIFAR-100} \citep{krizhevsky2009learning} that
\begin{enumerate}
   \item we can train state-of-the-art classifiers with a mixture of sufficiently-labeled and fully-labeled data, and
   \item only a small set of fully-labeled data is needed if given enough sufficiently-labeled data. 
\end{enumerate}
In particular, we show that the reduction on the need for fully-labeled data in practice is even more drastic that what is guaranteed by Theorem~\ref{th2}:
A single randomly-chosen full label per class was all the learning algorithm needed in all cases tested.
This means that state-of-the-art classifiers can be trained with almost only sufficiently-labeled data in practice.

\subsection{The Basic Settings}
The settings for these experiments are as follows.
We trained a LeNet-5 \citep{lecun1998gradient} for MNIST and a ResNet-18 \citep{he2016deep} for Fashion-MNIST, SVHN, and CIFAR-10.
To make sure that the model satisfied the assumption in Section~\ref{sec1}, the activation vector of the nonlinearity before the final linear layer was always normalized to \(1\) by (elementwise) dividing itself by its norm. 
This normalization did not affect performance.
The optimizer used was stochastic gradient descent with batch size \(128\).
Each hidden module (or full model, in the cases where only full labels were used) was trained with step size \(0.1\), \(0.01\), and \(0.001\), for \(200\), \(100\), and \(50\) epochs, respectively.
\rch{For CIFAR-100 in ``online'' mode, the hidden module was trained for 1000, 800, and 400 epochs with the said step sizes.}
If needed, each output layer was trained with step size \(0.1\) for \(50\) epochs.
For MNIST, the data was preprocessed by training set sample mean subtraction and then division by training set sample standard deviation.
For Fashion-MNIST, SVHN, CIFAR-10, \rch{and CIFAR-100}, the data was randomly cropped and flipped after the said normalization procedure.
For a training set of size \(50\rch{\text{k}}/100\rch{\text{k}}/200\rch{\text{k}}/60\rch{\text{k}}/120\rch{\text{k}}/240\rch{\text{k}}/73,257/146,514/293,028/\rch{400\text{k}/800\text{k}}\), we randomly selected \(5\rch{\text{k}}/10\rch{\text{k}}/20\rch{\text{k}}/5\rch{\text{k}}/10\rch{\text{k}}/20\rch{\text{k}}/5\rch{\text{k}}/10\rch{\text{k}}/20\rch{\text{k}/40\text{k}/80\text{k}}\) training examples to form the validation set.
The validation set was used for tuning hyperparameters and for determining the best model to save during each training session.
For a training set of size \(10\), no validation data was used and the optimal model was chosen based on the convergence of the loss function value on training data.
For SVHN, we did not use the ``additional'' training images.
For all datasets, we report performance on the standard test sets.
\rch{And standard supervised learning (training using only and all of the available fully-labeled data), which is the strongest baseline there is, will be used for comparison.}
We did not notice any of the proposed \(\ell_1\)'s significantly outperforming the others.
In all cases, we report performance using NCS.

\subsection{Obtaining Sufficient Labels}
\label{par1}

All of these curated datasets we selected are fully-labeled, and the sufficiently-labeled datasets used for the following experiments were derived from fully-labeled data. 
For example, to create a sufficiently-labeled training set with size \(100\rch{\text{k}}\), we randomly sampled \(100\rch{\text{k}}\) pairs of examples from the original fully-labeled training set.
And for each pair, a sufficient label of \(1\) or \(0\) was generated based on their original full labels. 
We discarded some pairs that do not contain useful information.
Specifically, suppose the input examples in the original dataset are \(\{x_i\}_{i=1}^n\), then the sampled set of example pairs is \(\left\{x_{1_i}, x_{2_i}\right\}_{i=1}^{n_1}\) with \(1_i < 2_i\) and \(\left(1_i, 2_i\right)\neq\left(1_j, 2_j\right), \forall i\neq j\).
Such a sufficiently-labeled training set covers \(n_1/(n(n-1)/2)\) of all possible informative example pairs from the original training set.

To approximate the performance upper bound for training with sufficiently-labeled data using these curated datasets, we used an online random sampling strategy (``online'').
Namely, in each training epoch, we iterated the original fully-labeled dataset in batches of size \(128\) and for each batch of fully-labeled data, we converted it into a set of sufficiently-labeled pairs (by taking pairwise combination, with uninformative pairs discarded as before) and used them to compute a model update step. 
And since the dataset was randomly shuffled at each epoch, we can potentially exhaust all sufficiently-labeled pairs given enough training epochs. 
In ``online'', the size of the validation set was chosen based on the size of the original fully-labeled dataset using the rules above.

\subsection{Results}
From Table~\ref{table1}, we see that state-of-the-art classifiers can be trained using almost only sufficiently-labeled data.
Indeed, a single randomly-chosen fully-labeled example from each class \rch{--- an amount that is trivial to obtain in practice ---} sufficed when learning with sufficient labels in all cases.
This validates Theorem~\ref{th1} and Theorem~\ref{th2}.
Also, to achieve similar test performance, learning with a mixture of sufficiently-labeled and fully-labeled data requires a training set with size similar to that of the training set required by learning with only fully-labeled data.
This verifies the earlier claim that learning with sufficient labels enjoys similar sample complexity as learning with purely full labels. 

\rch{
   Note that the sufficiently-labeled datasets we used were converted from datasets curated for the purpose of standard supervised training with fully-labeled data, and these ``repurposed'' datasets can be suboptimal for learning with sufficient labels due to the number of different-class pairs significantly outnumbering that of the same-class pairs.
   Take CIFAR-100 as an example, the fully-labeled dataset has \(100\) classes with the data population carefully chosen such that each class is represented by precisely \(1\%\) of the population, then if we convert the dataset into a sufficiently-labeled one by exhausting all pairwise combinations, examples with sufficient label \(0\) will dominate the resulting sufficiently-labeled population (roughly \(99\%\) sufficient labels will be \(0\)).
   In online mode, we found that good performance can still be obtained after prolonged training, which is perhaps because the model will eventually see enough same-class pairs in terms of absolute example count.
   In the modes where we trained with a predetermined number of sufficiently-labeled examples, we controlled the ratio of same-class to different-class pairs to make sure that there are enough same-class pairs in the training set. 
   Specifically, when we created the sufficiently-labeled training set by sampling examples from the original fully-labeled set, the sampling was performed in batches.
   In each batch, the sampler randomly selected $M$ unique classes, where $M$ was much smaller than the total number of classes.
   Then, for this batch, it randomly sampled data pairs only from these $M$ classes. 
   The need for keeping enough same-class pairs in the dataset is perhaps why we needed more data to match the performance of online or training with fully-labeled data especially on datasets with many classes such as CIFAR-100.
}

\rch{
   In general, a class distribution calibrated for fully-labeled learning may not be ideal for sufficiently-labeled learning.
   And we expect the performance of learning with sufficient labels to improve with tailored datasets.
}

\section{Conclusion}
\label{conclusions}

We proposed a novel form of labeled training data for classification that we called ``sufficiently-labeled'' data.
Sufficiently-labeled data can be obtained directly from annotators and it is easier to collect sufficiently-labeled data than fully-labeled data.
Sufficiently-labeled data can also serve as a layer of encryption on user information that cannot be fully broken.

A training algorithm that can learn from a mixture of sufficiently-labeled and fully-labeled data was proposed and analyzed. 
We proved that it can learn performant models and empirically showed that it shares a similar sample complexity as ERM using purely fully-labeled data.
We also showed that having more sufficient labels means needing less full labels.
And, in practice, we demonstrated that state-of-the-art classifiers can be trained with as few as a single random full label from each class when given enough sufficient labels.


\acks{ This work was supported by DARPA (FA9453-18-1-0039) and ONR (N00014-18-1-2306). }


\appendix
\onecolumn
\section*{Appendix A. Proof of Theorem~\ref{th1}}  
\begin{proof}
   \begin{align}
         &R(f_2\circ F_1, X, Y)\\\nonumber 
         &= E_{X, Y}\ell(f_2\circ F_1(X), Y)\\
         &= E_Y E_{X|Y} \ell(f_2\circ F_1(X), Y)\\
         &= \alpha E_{X|Y=+}\ell_+(f_2\circ F_1(X)) + (1 - \alpha) E_{X|Y=-}\ell_-(f_2\circ F_1(X))\\
         &= \alpha\int\ell_+(f_2\circ F_1(x))p_{X|Y=+}(x)dx + (1 - \alpha) \int\ell_-(f_2\circ F_1(x))p_{X|Y=-}(x)dx\\
         &= \int\int\left[\alpha\ell_+(f_2\circ F_1(x)) + (1 - \alpha) \ell_-(f_2\circ F_1(x^\prime))\right]p_{X|Y=+}(x)p_{X^\prime|Y^\prime=-}(x^\prime)dxdx^\prime\\
         &= \int\int\left[\alpha\ell_+(f_2\circ F_1(x)) + (1 - \alpha) \ell_-(f_2\circ F_1(x^\prime))\right]p_{X, X^\prime|Y=+, Y^\prime=-}(x, x^\prime)dxdx^\prime\\
         &= E_{X, X^\prime|Y=+, Y^\prime=-}\left[\alpha\ell_+(f_2\circ F_1(X)) + (1 - \alpha) \ell_-(f_2\circ F_1(X^\prime))\right],
   \end{align}
   where \(\ell_+(u) = - u\) and \(\ell_-(u) = + u\), and we have used the conditional independence of \(X, X^\prime\).
   \begin{align}
         &R(f_2\circ F_1, X, Y)\\\nonumber 
         &= E_{X, X^\prime|Y=+, Y^\prime=-}\left[-\left\langle w, \alpha\phi\circ F_1(X) - (1 - \alpha)\phi\circ F_1(X^\prime) \right\rangle\right]\\
         &= E_{X, X^\prime|Y=+, Y^\prime=-}\left[\|w\|\cos\theta(w, F_1, X, X^\prime) \left(-\left\|\alpha\phi\circ F_1(X) - (1 - \alpha)\phi\circ F_1(X^\prime)\right\|\right)\right], 
   \end{align}
   where \(\theta(w, F_1, X, X^\prime)\) is the angle between \(w\) and \(\alpha\phi\circ F_1(X) - (1 - \alpha)\phi\circ F_1(X^\prime)\) defined through the inner product when 
   \begin{equation}
      \|w\| > 0 \text{ and } \left\|\alpha\phi\circ F_1(X) - (1 - \alpha)\phi\circ F_1(X^\prime)\right\| > 0
   \end{equation}
    and can be assigned any arbitrary value when either norm is \(0\) since this leaves the expectation unchanged.

   Note that \(F_1\in\mathbb{S}\) if and only if 
   \begin{align}
      \Pr\left(\phi\circ F_1(X) = u_+|Y=+\right) = \Pr\left(\phi\circ F_1(X^\prime) = u_-|Y^\prime=-\right) = 1, 
   \end{align}
   for some constant vectors \(u_+ \neq \lambda u_-, \forall\lambda\geq 0\).
   The ``if'' direction is obvious.
   To prove the other direction, note that by conditional independence, if \(u\in\suppc{\phi\circ F_1(X)}\) given \(Y = +\) (or \(-\)) and \(v\in\suppc{\phi\circ F_1\left(X^\prime\right)}\) given \(Y^\prime = Y\), then \((u, v)\in\suppc{\left(\phi\circ F_1(X), \phi\circ F_1\left(X^\prime\right)\right)}\) given \(Y = Y^\prime\).
   Now, suppose \(u_1, u_2\) with \(u_1\neq u_2\) are both in \(\suppc{\phi\circ F_1(X)}\) given \(Y = + \) (or \(-\)) and let \(v\in\suppc{\phi\circ F_1\left(X^\prime\right)}\) given \(Y=Y^\prime\), then \(\phi\circ F_1(X) = \phi\circ F_1\left(X^\prime\right)\) cannot be true w.p. \(1\) given \(Y=Y^\prime\) since both \(\left(u_1, v\right)\) and \(\left(u_2, v\right)\) are in the support and \(v\) cannot be equal to \(u_1\) and \(u_2\) simultaneously.
   Therefore, \(\phi\circ F_1(X) = \phi\circ F_1\left(X^\prime\right)\) w.p. \(1\) given \(Y=Y^\prime\) implies that \(\phi\circ F_1(X)\) is degenerate given \(Y\).
   To prove \(u_+\neq \lambda u_-, \forall\lambda\geq 0\), note that since \(\left\|u_+\right\| = \left\|u_-\right\| > 0\), the only possible \(\lambda\) is \(1\). 
   But if \(u_+ = u_-\), then given \(Y \neq Y^\prime\), \(\phi\circ F_1(X) = \phi\circ F_1\left(X^\prime\right)\) w.p. \(1\), a contradiction. 

   Now, to prove
   \begin{equation}
      \mathbb{F}_1^\star\supseteq\mathbb{D}\cap\mathbb{S},
   \end{equation}
   it suffices to show that if 
   \begin{align}
         &F_1^\star\in\och{\argmin_{F_1\in\mathbb{F}_1}E_{X, X^\prime|Y = +, Y^\prime = -} \left(-\left\|\phi\circ F_1(X) - \phi\circ F_1(X^\prime)\right\|^2\right)}\label{eq9}\\ 
         & \och{= \argmin_{F_1\in\mathbb{F}_1}E_{X, X^\prime|Y = +, Y^\prime = -}\left(\left\langle\phi\circ F_1(X), \phi\circ F_1(X^\prime)\right\rangle\right)},\text{ and}\nonumber\\ 
         &\Pr\left(\phi\circ F_1^\star(X) = u_+^\star|Y=+\right) = \Pr\left(\phi\circ F_1^\star(X^\prime) = u_-^\star|Y^\prime=-\right) = 1, \label{eq6} 
   \end{align}
   for some \(u_+^\star\neq \lambda u_-^\star, \forall \lambda\geq 0\), then
   \begin{equation}
      \min_{f_2\in\mathbb{F}_2} R(f_2\circ F_1^\star, X, Y)\leq \min_{f_2\in\mathbb{F}_2}R(f_2\circ F_1, X, Y), \forall F_1\in\mathbb{F}_1.\label{eq12}
   \end{equation} 

To this end, note that Eq.~\ref{eq6} implies \(\alpha\phi\circ F_1^\star(X) - (1 - \alpha)\phi\circ F_1^\star(X^\prime)\) is a nonzero constant vector w.p. \(1\) given \(Y = +, Y^\prime = -\). 
Therefore, given \(Y=+, Y^\prime = -\), for any \(\och{1/r\geq} a > 0\), there exists \(w^\star\) with \(\|w^\star\| = a\) and \(\cos\theta(w^\star, F_1^\star, X, X^\prime) = 1\) w.p. \(1\).\footnote{In fact, by Cauchy-Schwarz inequality, \(a\left(\alpha u_+^\star - (1 - \alpha)u_-^\star\right)/\left\|\alpha u_+^\star - (1 - \alpha)u_-^\star\right\|\) is the unique \(w^\star\).}

For any \(F_1\in\mathbb{F}_1\), let an \(f_2\in\rch{\argmin_{f_2\in\mathbb{F}_2}}R(f_2\circ F_1, X, Y)\) be parameterized by \(w\).
If \(\|w\|>0\), we can then find \(w^\star\) such that \(\|w^\star\| = \|w\|\) and \(\cos\theta(w^\star, F_1^\star, X, X^\prime) = 1\) w.p. \(1\).
On the other hand, if \(\|w\| = 0\), find \(w^\star\) such that \(\|w^\star\| = a\) and \(\cos\theta(w^\star, F_1^\star, X, X^\prime) = 1\) w.p. \(1\) for some \(\och{1/r\geq} a>0\). 
In both cases, we have
\begin{align}
      &\min_{f_2\in\mathbb{F}_2}R(f_2\circ F_1, X, Y)\\\nonumber
      &=  E_{X, X^\prime|Y=+, Y^\prime=-}\left[\|w\|\cos\theta(w, F_1, X, X^\prime) \left(-\left\|\alpha\phi\circ F_1(X) - (1 - \alpha)\phi\circ F_1(X^\prime)\right\|\right)\right]\\
      &\geq E_{X, X^\prime|Y=+, Y^\prime=-}\left[\|w^\star\| \left(-\left\|\alpha\phi\circ F_1(X) - (1 - \alpha)\phi\circ F_1(X^\prime)\right\|\right)\right]\\
      &= \|w^\star\| E_{X, X^\prime|Y=+, Y^\prime=-}\left(-\left\|\alpha\phi\circ F_1(X) - (1 - \alpha)\phi\circ F_1(X^\prime)\right\|\right)\label{eq10}\\
      &\och{= \|w^\star\| E_{X, X^\prime|Y=+, Y^\prime=-}\left(-\sqrt{\left\|\alpha\phi\circ F_1(X) - (1 - \alpha)\phi\circ F_1(X^\prime)\right\|^2}\right)}\\
      &\och{\stackrel{\text{(Jensen's Inequality)}}{\geq} \|w^\star\| \left(-\sqrt{E_{X, X^\prime|Y=+, Y^\prime=-}\left\|\alpha\phi\circ F_1(X) - (1 - \alpha)\phi\circ F_1(X^\prime)\right\|^2}\right)}\label{eq13}\\
      &\och{= \|w^\star\| \left(-\sqrt{\left(\alpha^2 + (1 - \alpha)^2\right)r^2 - 2\alpha(1 - \alpha)E_{X, X^\prime|Y=+, Y^\prime=-}\left\langle\phi\circ F_1(X), \phi\circ F_1(X^\prime)\right\rangle}\right)}\\
      &\och{\stackrel{\text{(Eq.~\ref{eq9})}}{\geq} \|w^\star\| \left(-\sqrt{\left(\alpha^2 + (1 - \alpha)^2\right)r^2 - 2\alpha(1 - \alpha)E_{X, X^\prime|Y=+, Y^\prime=-}\left\langle\phi\circ F^\star_1(X), \phi\circ F^\star_1(X^\prime)\right\rangle}\right)}\\
      &\och{= \|w^\star\| \left(-\sqrt{E_{X, X^\prime|Y=+, Y^\prime=-}\left\|\alpha\phi\circ F^\star_1(X) - (1 - \alpha)\phi\circ F^\star_1(X^\prime)\right\|^2}\right)}\\
      &\och{\stackrel{\text{(Eq.~\ref{eq6})}}{=} \|w^\star\| \left(-\sqrt{\left\|\alpha u_+^\star - (1 - \alpha)u_-^\star\right\|^2}\right)}\\
      &\och{= \|w^\star\| E_{X, X^\prime|Y=+, Y^\prime=-}\left(-\left\|\alpha\phi\circ F^\star_1(X) - (1 - \alpha)\phi\circ F^\star_1(X^\prime)\right\|\right)}\label{eq11}\\
      &=  E_{X, X^\prime|Y=+, Y^\prime=-}\left[\|w^\star\|\cos\theta(w^\star, F_1^\star, X, X^\prime) \left(-\left\|\alpha\phi\circ F_1^\star(X) - (1 - \alpha)\phi\circ F_1^\star(X^\prime)\right\|\right)\right]\\
      &\geq \min_{f_2\in\mathbb{F}_2} R(f_2\circ F_1^\star, X, Y). 
\end{align}

On the other hand, to prove 
   \begin{equation}
      \mathbb{F}_1^\star\subseteq\mathbb{S}\cap\mathbb{D},
   \end{equation}
let an \(F_1\in\mathbb{F}_1^\star\) be given, we need to show 
\begin{equation}
   F_1\notin\stcomp{\mathbb{S}}\cup\stcomp{\mathbb{D}} = \left(\stcomp{\mathbb{S}}\cap\mathbb{D}\right)\cup\stcomp{\mathbb{D}}.
\end{equation}

 If \(F_1\in\stcomp{\mathbb{D}}\), there exists \(F_1^\star\in\mathbb{S}\) such that
\begin{align}
      &\och{E_{X, X^\prime|Y=+, Y^\prime=-}\left(-\left\|\phi\circ F_1(X) - \phi\circ F_1(X^\prime)\right\|^2\right)}\\\nonumber
      &\och{>E_{X, X^\prime|Y=+, Y^\prime=-}\left(-\left\|\phi\circ F_1^\star(X) - \phi\circ F_1^\star(X^\prime)\right\|^2\right)}.
\end{align}
Let an \(f_2\in\rch{\argmin_{f_2\in\mathbb{F}_2}}R(f_2\circ F_1, X, Y)\) be parameterized by \(w\).
We have shown that we can find \(w^\star\) such that \(\|w^\star\| = \|w\|\) (or \(\|w^\star\| = a\) for some \(\och{1/r \geq} a > 0\) if \(\|w\| = 0\)) and \(\cos\theta(w^\star, F_1^\star, X, X^\prime) = 1\) w.p. \(1\). 
Then we have
\begin{align}
      &\min_{f_2\in\mathbb{F}_2}R(f_2\circ F_1, X, Y)\\\nonumber
      &=  E_{X, X^\prime|Y=+, Y^\prime=-}\left[\|w\|\cos\theta(w, F_1, X, X^\prime) \left(-\left\|\alpha\phi\circ F_1(X) - (1 - \alpha)\phi\circ F_1(X^\prime)\right\|\right)\right]\\
      &\geq E_{X, X^\prime|Y=+, Y^\prime=-}\left[\|w^\star\| \left(-\left\|\alpha\phi\circ F_1(X) - (1 - \alpha)\phi\circ F_1(X^\prime)\right\|\right)\right]\\
      &= \|w^\star\| E_{X, X^\prime|Y=+, Y^\prime=-}\left(-\left\|\alpha\phi\circ F_1(X) - (1 - \alpha)\phi\circ F_1(X^\prime)\right\|\right)\\
      &\stackrel{\text{\och{(Similar to Eq.~\ref{eq10} - Eq.~\ref{eq11})}}}{>} \|w^\star\| E_{X, X^\prime|Y=+, Y^\prime=-}\left(-\left\|\alpha\phi\circ F_1^\star(X) - (1 - \alpha)\phi\circ F_1^\star(X^\prime)\right\|\right)\\
      &=  E_{X, X^\prime|Y=+, Y^\prime=-}\left[\|w^\star\|\cos\theta(w^\star, F_1^\star, X, X^\prime) \left(-\left\|\alpha\phi\circ F_1^\star(X) - (1 - \alpha)\phi\circ F_1^\star(X^\prime)\right\|\right)\right]\\
      &\geq \min_{f_2\in\mathbb{F}_2} R(f_2\circ F_1^\star, X, Y), 
\end{align}
contradicting \(F_1\in\mathbb{F}_1^\star\).

On the other hand, if \(F_1\in\stcomp{\mathbb{S}}\cap\mathbb{D}\), then let \(F_1^\star\in\mathbb{S}\cap\mathbb{D}\) and we have 
\begin{align}
      &\och{E_{X, X^\prime|Y=+, Y^\prime=-}\left(-\left\|\phi\circ F_1(X) - \phi\circ F_1(X^\prime)\right\|^2\right)}\\\nonumber
      &\och{=E_{X, X^\prime|Y=+, Y^\prime=-}\left(-\left\|\phi\circ F_1^\star(X) - \phi\circ F_1^\star(X^\prime)\right\|^2\right)}.
\end{align}
Let an \(f_2\in\rch{\argmin_{f_2\in\mathbb{F}_2}}R(f_2\circ F_1, X, Y)\) be parameterized by \(w\).
Again, we can find \(w^\star\) such that \(\|w^\star\| = \|w\|\) (or \(\|w^\star\| = a\) for some \(\och{1/r \geq } a > 0\) if \(\|w\| = 0\)) and \(\cos\theta(w^\star, F_1^\star, X, X^\prime) = 1\) w.p. \(1\). 

Suppose we can show that there does not exist \(w\) such that \(w\) is linearly dependent with \(\alpha\phi\circ F_1(X) - (1 - \alpha)\phi\circ F_1\left(X^\prime\right)\) w.p. \(1\) given \(Y = +, Y^\prime = -, \left\|\alpha\phi\circ F_1(X) - (1 - \alpha)\phi\circ F_1\left(X^\prime\right)\right\| > 0\), then  
\begin{align}
      &\min_{f_2\in\mathbb{F}_2}R(f_2\circ F_1, X, Y)\\\nonumber
      &=  E_{X, X^\prime|Y=+, Y^\prime=-}\left[-\left\langle w, \alpha\phi\circ F_1(X) - (1 - \alpha)\phi\circ F_1(X^\prime)\right\rangle\right]\\
      &> E_{X, X^\prime|Y=+, Y^\prime=-}\left[\|w\| \left(-\left\|\alpha\phi\circ F_1(X) - (1 - \alpha)\phi\circ F_1(X^\prime)\right\|\right)\right]\\
      &= \|w^\star\| E_{X, X^\prime|Y=+, Y^\prime=-}\left(-\left\|\alpha\phi\circ F_1(X) - (1 - \alpha)\phi\circ F_1(X^\prime)\right\|\right)\\
      &\stackrel{\text{\och{(Similar to Eq.~\ref{eq10} - Eq.~\ref{eq11})}}}{\och{\geq}} \|w^\star\| E_{X, X^\prime|Y=+, Y^\prime=-}\left(-\left\|\alpha\phi\circ F_1^\star(X) - (1 - \alpha)\phi\circ F_1^\star(X^\prime)\right\|\right)\\
      &=  E_{X, X^\prime|Y=+, Y^\prime=-}\left[\|w^\star\|\cos\theta(w^\star, F_1^\star, X, X^\prime) \left(-\left\|\alpha\phi\circ F_1^\star(X) - (1 - \alpha)\phi\circ F_1^\star(X^\prime)\right\|\right)\right]\\
      &\geq \min_{f_2\in\mathbb{F}_2} R(f_2\circ F_1^\star, X, Y), 
\end{align}
which, again, contradicts \(F_1\in\mathbb{F}_1^\star\).

Finally, to show that there does not exist \(w\) such that \(w\) is linearly dependent with \(\alpha\phi\circ F_1(X) - (1 - \alpha)\phi\circ F_1\left(X^\prime\right)\) w.p. \(1\) given \(Y = +, Y^\prime = -, \left\|\alpha\phi\circ F_1(X) - (1 - \alpha)\phi\circ F_1\left(X^\prime\right)\right\| > 0\), we use the fact that \(F_1\notin\mathbb{S}\). 
First, note that for any \(w\), \(\alpha\phi\circ F_1(X) - (1 - \alpha)\phi\circ F_1\left(X^\prime\right)\) being linearly dependent with \(w\) w.p. \(1\) given \(Y = +, Y^\prime = -, \left\|\alpha\phi\circ F_1(X) - (1 - \alpha)\phi\circ F_1\left(X^\prime\right)\right\| > 0\) implies that the support of \(\alpha\phi\circ F_1(X) - (1 - \alpha)\phi\circ F_1\left(X^\prime\right)\) given \(Y=+, Y^\prime = -\) is a subset of \(\{\lambda w | \lambda\in\mathbb{R}\}\). 
By conditional independence of \(X, X^\prime\) and the assumption that \(\|\phi(u)\| = r, \forall u\), any \(u\in\suppc{\alpha\phi\circ F_1(X)}\) given \(Y = +\) and \(v\in\suppc{(1 - \alpha)\phi\circ F_1\left(X^\prime\right)}\) given \(Y^\prime = -\) have \(u - v\in\suppc{\left(\alpha\phi\circ F_1(X) - (1 - \alpha) \phi\circ F_1\left(X^\prime\right)\right)}\) given \(Y = +, Y^\prime = -\) and \(\|u\| = \alpha/(1 - \alpha)\|v\|\).
Meanwhile, for a given \(v\) 
\begin{equation}
   u - v = \lambda w\Rightarrow \|v + \lambda w\| = \|u\| = \frac{\alpha}{1 - \alpha}\|v\|
\end{equation}
can be true for at most two distinct values of \(\lambda\).
This implies 
\begin{equation}
   \label{eq7}
   \left|\suppc{\phi\circ F_1(X)} \text{ given } Y = + \right|\leq 2.
\end{equation}
Similarly, we have 
\begin{equation}
   \label{eq8}
   \left|\suppc{\phi\circ F_1\left(X^\prime\right)} \text{ given } Y^\prime = - \right|\leq 2.
\end{equation}
Eq.~\ref{eq7} and \ref{eq8} together gives 
\begin{eqnarray}
   \left|\suppc{\phi\circ F_1(X)}\right|\leq 4, 
\end{eqnarray}
contradicting \(F_1\notin\mathbb{S}\) due to the assumption that \(\left|\suppc{\phi\circ F_1(X)}\right|> 4, \forall F_1\notin\mathbb{S}\).
\end{proof}

\section*{Appendix B. Proof of Theorem~\ref{th2}}
\begin{proof}
For any \(F_1\), define 
\begin{equation}
   \ell\circ\mathbb{F}_{2, A}\circ F_1 = \{\ell\circ f_2\circ F_1| f_2\in\mathbb{F}_{2, A}\}.
\end{equation}
Then let \rch{\(\mathcal{R}_n\)} denote the \rch{Rademacher} complexity, we have \citep{bartlett2002rademacher},
\begin{equation}
   \rch{\mathcal{R}_n}(\ell\circ\mathbb{F}_{2, A}\circ F_1)\leq \frac{Ar}{\sqrt{n}}.
\end{equation}

Note that 
\begin{equation}
   |\ell(f_2\circ F_1(x), y)| \leq \|w\|r \leq Ar, \forall f_2\in\mathbb{F}_{2, A}.
\end{equation} 

Let a true risk value \(\gamma<0\) and any probability \(\delta\) be given, then, with probability at least \(1 - \delta\), we have \citep{bartlett2002rademacher,shalev2014understanding}  
\begin{equation}
   \sup_{\hat{f}_2\in\rchnew{\hat{\mathbb{F}}_{2, F_1, \gamma}^\star}}R\left(\hat{f}_2\circ F_1, X, Y\right) - \gamma\leq \rch{2\frac{t\left(F_1\right)r}{\sqrt{n_2}}} + 5t\left(F_1\right)r\sqrt{\frac{2\ln\left(8/\delta\right)}{n_2}},
\end{equation}
where \(t(F_1)\) is the function that gives the smallest \(A\) such that for all \(f_2\in\mathbb{F}_{2, F_1, \gamma}\), we have \(\|w\|\leq A\).  

Now, to prove \(\mathbb{S}\cap\mathbb{D} \subseteq \argmin_{F_1\in\mathbb{F}_1}t(F_1)\), let \(F_1^\star\in\mathbb{S}\cap\mathbb{D}\) and an arbitrary eligible \(F_1\) be given.
We have seen in the proof of Theorem~\ref{th1} that \(\left\|\alpha\phi\circ F_1^\star(X) - (1 - \alpha)\phi\circ F_1^\star(X^\prime) \right\|\neq 0\) w.p. \(1\) given \(Y=+, Y^\prime = -\).
Therefore, \(E_{X, X^\prime|Y=+, Y^\prime=-}\left(-\left\|\alpha\phi\circ F_1^\star(X) - (1 - \alpha)\phi\circ F_1^\star(X^\prime) \right\|\right) \neq 0\). 
Also, define \(\theta\left(w, F_1, X, X^\prime\right)\) as in the proof of Theorem~\ref{th1}.
Let a particular \(f_2\) be the \(f_2\) with the smallest \(\|w\|\) among all \(f_2\) such that \(f_2\circ F_1\) attains true risk \(\gamma\).
Suppose this \(f_2\) is parameterized by \(w\). 
Evidently, \(t(F_1) = \|w\|\).
And we have
\begin{align}
   \gamma &= R(f_2\circ F_1, X, Y)\\
   &= E_{X, X^\prime|Y=+, Y^\prime=-}\|w\|\cos\theta(w, F_1, X, X^\prime)\left(-\left\|\alpha\phi\circ F_1(X) - (1 - \alpha)\phi\circ F_1(X^\prime) \right\|\right)\\
   &\geq \|w\|E_{X, X^\prime|Y=+, Y^\prime=-}\left(-\left\|\alpha\phi\circ F_1(X) - (1 - \alpha)\phi\circ F_1(X^\prime) \right\|\right)\\
   &\och{\stackrel{\text{(Jensen's Inequality)}}{\geq} \|w\|\left(-\sqrt{E_{X, X^\prime|Y=+, Y^\prime=-}\left\|\alpha\phi\circ F_1(X) - (1 - \alpha)\phi\circ F_1(X^\prime) \right\|^2}\right)}\\
   &\och{= \|w\|\sqrt{\frac{E_{X, X^\prime|Y=+, Y^\prime=-}\left\|\alpha\phi\circ F_1(X) - (1 - \alpha)\phi\circ F_1(X^\prime) \right\|^2}{E_{X, X^\prime|Y=+, Y^\prime=-}\left\|\alpha\phi\circ F_1^\star(X) - (1 - \alpha)\phi\circ F_1^\star(X^\prime) \right\|^2}}}\\
   &\och{\qquad\qquad \left(-\sqrt{E_{X, X^\prime|Y=+, Y^\prime=-}\left\|\alpha\phi\circ F_1^\star(X) - (1 - \alpha)\phi\circ F_1^\star(X^\prime) \right\|^2}\right)}.
\end{align}
Using an argument from the proof of Theorem~\ref{th1} \rch{(paragraph following Eq.~\ref{eq12})}, \(F_1^\star\in\mathbb{S}\) implies the existence of an \(w^\star\) such that 
\begin{equation}
   \rch{\|w^\star\| = \|w\| \sqrt{\frac{E_{X, X^\prime|Y=+, Y^\prime=-}\left\|\alpha\phi\circ F_1(X) - (1 - \alpha)\phi\circ F_1(X^\prime) \right\|^2}{E_{X, X^\prime|Y=+, Y^\prime=-}\left\|\alpha\phi\circ F_1^\star(X) - (1 - \alpha)\phi\circ F_1^\star(X^\prime) \right\|^2}}\leq\|w\|}
\end{equation}
and \(\cos\theta(w^\star, F_1^\star, X, X^\prime) = 1\) w.p. \(1\), \rch{where the inequality is due to Eq.~\ref{eq9}}.
Then, \och{with derivation same as Eq.~\ref{eq13} - Eq.~\ref{eq11}}, we have
\begin{align}
   \gamma &\geq \|w^\star\| E_{X, X^\prime|Y=+, Y^\prime=-} \cos\theta(w^\star, F_1^\star, X, X^\prime)\left(-\left\|\alpha\phi\circ F_1^\star(X) - (1 - \alpha)\phi\circ F_1^\star(X^\prime) \right\|\right)\\
   &\geq\min_{f_2\in\mathbb{F}_{2, \|w^\star\|}}R(f_2\circ F_1^\star, X, Y),
\end{align}
which, combined with \(F_1^\star\in\mathbb{D}\) and the fact that another \(w^{\star\star}\) can be found (by, e.g., multiplying \(w^\star\) with a scalar) such that it parameterizes an \(f_2^{\star\star}\) with \(f_2^{\star\star}\circ F_1^\star\) attaining risk \(\gamma\) and \(\|w^{\star\star}\|\leq\|w^\star\|\), indicates 
\begin{equation}  
   t(F_1^\star) \leq \|w^\star\| \leq \|w\| = t(F_1),
\end{equation}
proving \(\mathbb{S}\cap\mathbb{D}\subseteq\argmin_{F_1\in\mathbb{F}_1} t(F_1)\).

On the other hand, assume \(F_1\notin\mathbb{S}\cap\mathbb{D}\) but \(F_1\in\argmin_{F_1\in\mathbb{F}_1}t(F_1)\).
Suppose a particular \(f_2\) is the \(f_2\) with the smallest \(\|w\|\) among all \(f_2\) such that \(f_2\circ F_1\) attains \(\gamma\) true risk.
Let this particular \(f_2\) be parameterized by \(w\) and we have \(t(F_1) = \|w\|\).
Let \(F_1^\star\in\mathbb{S}\cap\mathbb{D}\) be given.

If \(F_1\notin\mathbb{D}\), then using a similar argument as above, we can find \(w^\star\) with \(\|w^\star\|<\|w\|\) such that \(f_2^\star\circ F_1^\star\) attains at most \(\gamma\) true risk and therefore \(t(F_1^\star)\leq\|w^\star\|<\|w\| = t(F_1)\), contradicting \(F_1\in\argmin_{F_1\in\mathbb{F}_1}t(F_1)\).

If \(F_1\in\mathbb{D}\) but \(F_1\notin\mathbb{S}\), then we have shown in the proof of Theorem~\ref{th1} that there exists \(w^\star\) such that \(\|w^\star\| = \|w\|, \och{\cos\theta(w^\star, F_1^\star, X, X^\prime) = 1,}\) but \(R\left(f_2^\star\circ F_1^\star, X, Y\right) < R\left(f_2\circ F_1, X, Y\right) = \gamma\), indicating that there exists \(w^{\star\star}\) parameterizing an \(f_2^{\star\star}\) with \(R\left(f_2^{\star\star}\circ F_1^\star, X, Y\right) = \gamma\) but \(\|w^{\star\star}\| < \|w^\star\| = \|w\|\) (one such \(w^{\star\star}\) is given by \(\omega w^\star\) for some \(\omega \in [0, 1)\)).
Thus, \(t(F_1^\star)\leq\|w^{\star\star}\|<\|w\|=t(F_1)\), again contradicting \(F_1\in\argmin_{F_1\in\mathbb{F}_1}t(F_1)\). 

\end{proof}

\vskip 0.2in
\bibliography{main}

\end{document}